\documentclass{article}

\usepackage[nonatbib,preprint]{neurips_2020}

\usepackage[utf8]{inputenc} 
\usepackage[T1]{fontenc}    
\usepackage{hyperref}       
\usepackage{url}            
\usepackage{booktabs}       
\usepackage{amsfonts}       
\usepackage{nicefrac}       
\usepackage{microtype}      

\usepackage[round]{natbib}

\usepackage{microtype}
\usepackage{amsmath}
\usepackage{wrapfig}

\usepackage{graphicx}
\usepackage{subcaption}

\newcommand{\sizeTicket}[0]{|t|} 

\title{How many winning tickets are there in one DNN?}

\author{%
  Kathrin Grosse, Michael Backes \\
  CISPA Helmholtz Center for Information Security\\
  \texttt{kathrin.grosse@cispa.saarland} \\
} 

\begin{document}

\maketitle


\begin{abstract}
The recent \emph{lottery ticket hypothesis} proposes that there is one sub-network that matches the accuracy of the original network when trained in isolation.
We show that instead each network contains 
several winning tickets, even if the initial weights are fixed. 
The resulting winning sub-networks are not instances of the same network under weight space symmetry, and show no overlap or correlation significantly larger than expected by chance.
If randomness during training is decreased, overlaps higher than chance occur, even if the networks are trained on different tasks.
We conclude that there is rather a distribution over capable sub-networks, as opposed to a single winning ticket.
\end{abstract}

\section{Introduction}
Good performance in deep learning often requires long training time to find good values for often millions of parameters~\citep{huang2017densely}.
Obtaining smaller networks can be desirable to save energy or time during prediction~\citep{han2015learning,luo2017thinet,liu2018rethinking,crowley2018pruning}.
Often, such pruned and thus smaller networks do not perform worse than their larger counterparts.

A recent trend proposed to prune during training time~\citep{frankle2018lottery,gomez2018targeted,desai2019evaluating,morcos2019one}, yielding an iterative process.
The network is trained, pruned, and training restarted with a smaller network.
This procedure is repeated several times, with more and more weights being removed.
Many approaches rely on the idea of a winning sub-network that emerges in this process,
a so called \emph{winning ticket}.
\citet{frankle2018lottery}
brought forward the original hypothesis introducing winning tickets. More concretely, the authors state  
\begin{itemize}
\item[] \textbf{The Lottery Ticket Hypothesis.} A randomly-initialized, dense neural network contains a sub-network that is initialized such that---when trained in isolation---it can match the test accuracy of the
original network after training for at most the same number of iterations.
\end{itemize}
\citet{anonymous2020the} state the possibility that there could be several such sub-networks or winning tickets.
The authors measure the Hamming distance between tickets of ResNet while varying the epochs pruning is applied and the epochs the weights are taken from when restarting training.
We take a slightly different approach and fix the initial weights for a task, and do not fix randomness during training.
Our experiments show that a new, different winning ticket emerges from each training run.

Orthogonally, \citet{morcos2019one} or \citet{desai2019evaluating}  
study whether winning tickets transfer across tasks or datasets.
We take the inverse step, and investigate how tickets 
overlap for different tasks when using the same initial weights and randomness is restricted during training. 

More concretely, our contributions are as follows. 
\textbf{First}, we show that even when using the same initial weights, there is a new, different winning ticket after each training run, given randomness is not fixed. \textbf{Second,}
the overlap of different tickets as expected by chance. 
There are no unexpected shared structures, and rank correlations between initial and final weights are scattered around zero.
\textbf{Third}, the resulting networks are inherently different, and not an artifact of weight-space symmetry.
\textbf{Fourth}, when randomness is partially fixed, overlap beyond chance occurs, even between networks trained on different data-sets. 
\textbf{Fifth}, the accuracies of the tickets is slightly higher when randomness is slightly fixed than with no randomness fixed.
\textbf{Sixth}, given the existence of many different winning tickets, we conclude that there is rather a distribution over capable sub-networks than a single winner for learning a particular task. A distribution over capable sub-networks, in particular when task independent, could lead to new heuristics improving initialization, an open research topic in deep learning \citep{zhang2019fixup}.

\subsection{Related Work}
Following the work of \citet{frankle2018lottery}, winning tickets have gotten a lot of attention. 
For example, \cite{gondara2020differentially} use tickets to obtain differentially private neural networks.
However, much effort focuses on tickets for very large neural networks or making training more efficient~\citep{vinyals2016matching,crowley2018pruning,gomez2018targeted,gaier2019weight}. 

As \cite{achille2017critical} showed that early training stages are important, \cite{you2019drawing} presented that evidence that winning tickets emerge early in training, too.
Generally, research sped up the process of finding winning tickets~\citep{morcos2019one,anonymous2020the}. Effort was also made to find tickets for very large networks~\citep{frankle2019lottery}.
We offer a different perspective on winning tickets, as we present evidence that there are many tickets, not only one. 

Orthogonal work by \cite{ramanujan2019s} shows that winning sub-networks can be distilled from a network without any training. In this case,  the structure is learned, not the weights.

\section{Experimental Set-Up}

 \begin{figure}[t]
  \begin{subfigure}{.54\textwidth}
  \includegraphics[width=\linewidth]{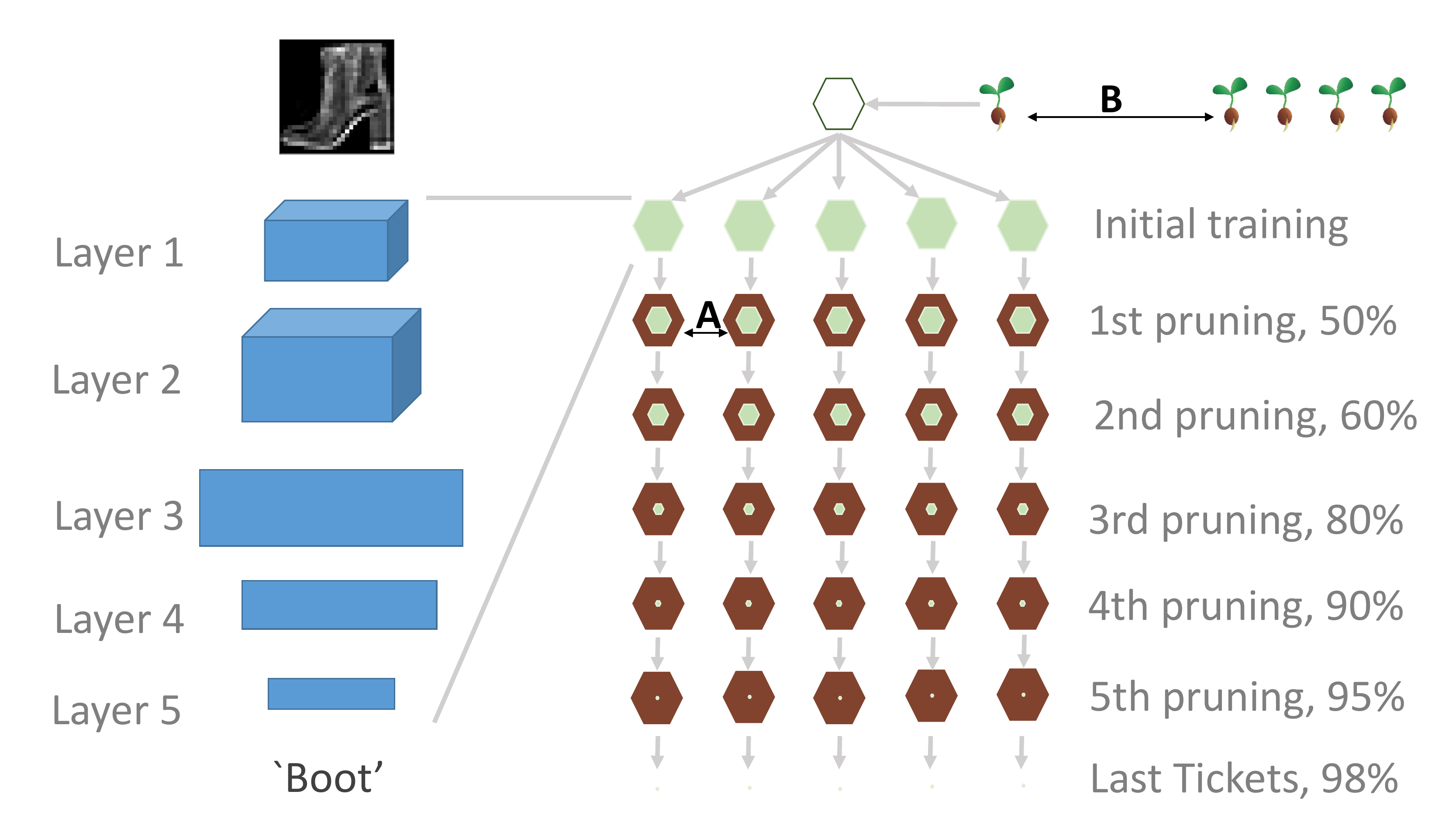}
    \caption{Experimental set-up, example using Fashion MNIST.}\label{fig:setUP}
  \end{subfigure}
    \begin{subfigure}{.49\textwidth}
  \includegraphics[width=\linewidth]{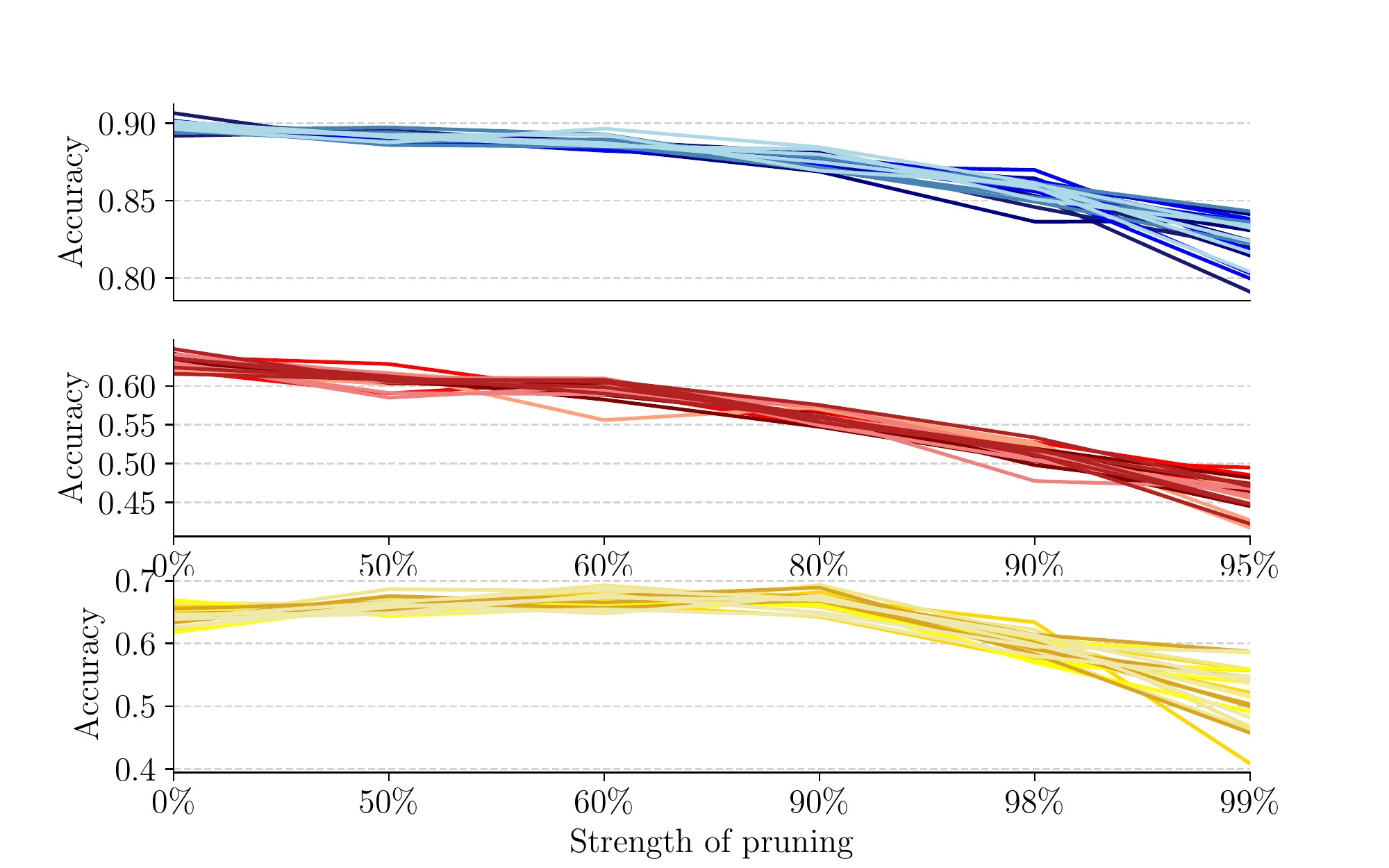}
  \caption{Accuracies of pruning steps.}\label{fig:Accs}
    \end{subfigure}
  \caption{(a): Experimental set-up. We fix an architecture and derive initial weights for a seed. These same weights are trained five times, computing tickets in an iterative retraining process. We compare the tickets obtained from one seed (\textbf{A}), across seeds (\textbf{B}), or across tasks (not visualized). (b): Accuracies for tasks, networks, and pruning steps. From top to bottom: Fashion MNIST, CIFAR10, ResNet18 on CIFAR10. Each line is one run, each color shade one seed. Plots are best seen in color.}
\end{figure}

We describe our experimental set-up, starting with the datasets, network architecture, and then describe the details of training and the used baselines.
The
Fashion-MNIST~\citep{xiao2017/online} and the CIFAR10~\citep{krizhevsky2009learning}
datasets are used. 
We apply a convolutional network, consisting in two convolutional layers (with 6 and 16 $5 \times 5$ filters each) and max-pooling. Two dense layers (with 120 and 84 units, respectively) follow before the softmax output. 
We further train a small ResNet18~\citep{he2016deep} to verify that our results hold independent of model size. We generally plot all layers of the small networks and chose randomly five layers of the ResNet (1,11,12,18, and 19) for visualizations. Additional layers can be plotted using the supplementary material.


Each network is trained for 15 epochs---we train few epochs as previous work shows that winning tickets emerge early in training \citep{achille2017critical,you2019drawing}.
We further obtain the winning tickets as stated in \citep{zhou2019deconstructing} (using the large final criterion). 
The code to reproduce all plots in this paper is available upon request.

The experimental set-up is visualized in Figure~\ref{fig:setUP}.
We choose five initial weight initializations (five random seeds). 
These five initial weights are kept, however we do not fix randomness in for example the sampling of the batches, and the gradients. 
On each of the five initial weights, we run five independent iterative pruning procedures.
Each of these five runs yields one final ticket in six pruning steps, as visualized in Figure~\ref{fig:setUP}. 
At each pruning step, we prune $50$\%, $60$\%, $80$\%, $90$\%, $95$\%, and $98$\% compared to the size of the initial/original weights. 
Relative to the kept weights, from first to second step, $20$\% more weights are pruned.
We prune slightly more weights for the ResNet, using percentages $50$\%, $60$\%, $90$\%, $98$\%, $99$\%, and $99.9$\%.
The reason is that accuracy is fairly stable for the percentages above, possibly due to the larger weight matrices or more overall weights due to skip connections (see Figure~\ref{fig:Accs}).
With the given setting, the small networks perform best in pruning steps one, and two, and afterwards decreases with stronger pruning. The ResNet performs best in pruning iteration one, two and three, then, accuracy decreases.
We also investigate cases of very small masks with decreased accuracy, as these networks still show good performance relative to their size.
For the remainder of the paper, 
we use the terms mask and ticket interchangeably.

\begin{wrapfigure}{r}{0.55\textwidth}
  \vspace{-10pt}
  \includegraphics[width=0.32\linewidth]{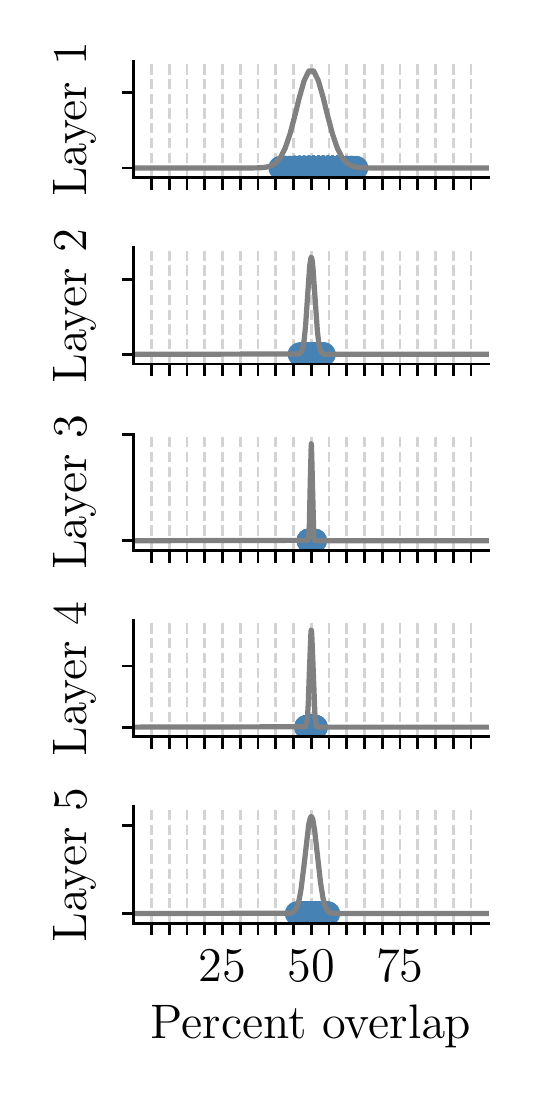}
  \includegraphics[width=0.32\linewidth]{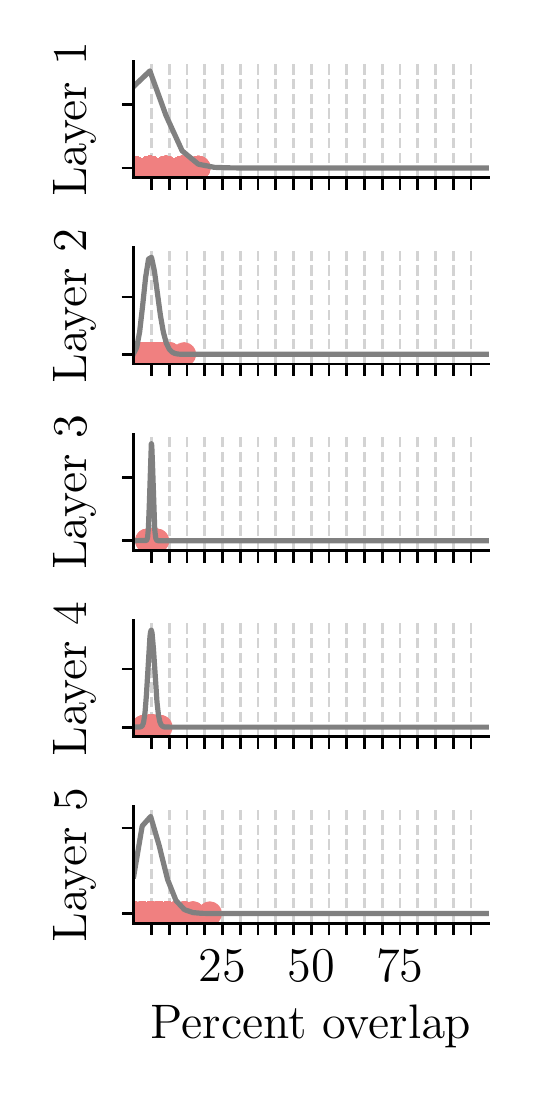}
  \includegraphics[width=0.32\linewidth]{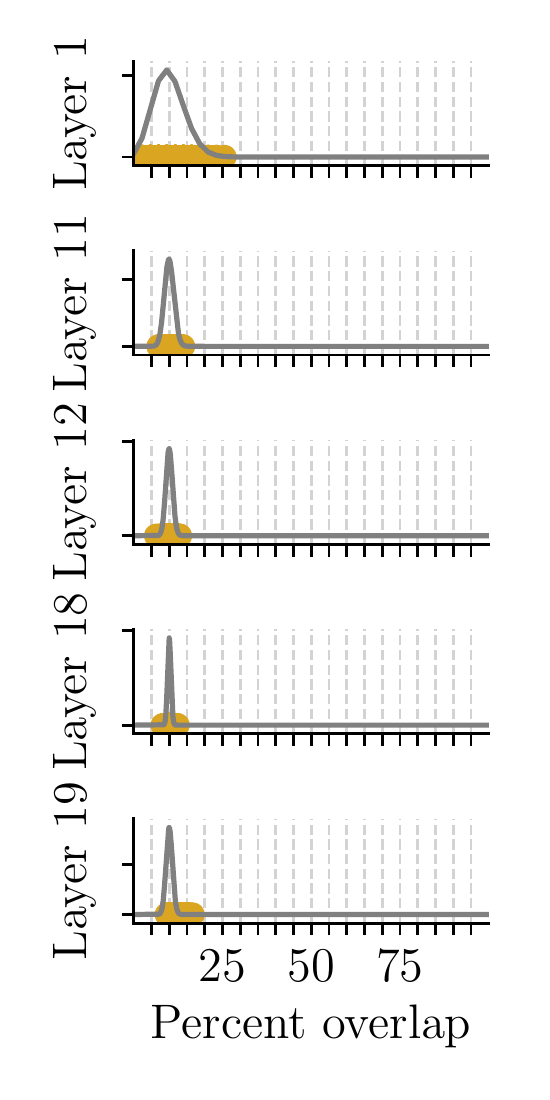}
  \caption{Overlap (in percent) between masks across initializations. From left to right: Fashion MNIST, first pruning step, CIFAR10, fifth pruning step, ResNet on CIFAR10, third pruning step. The gray curve is the expected overlap given the hyper geometric distribution.}
  \label{fig:overlapseedPrune}
  \vspace{-10pt}
\end{wrapfigure}

The amount of a random chance overlap between two masks is determined using the hyper-geometric distribution.
It specifies given a population (size of weight matrix  $m \times n$) and a number of objects with a particular feature (e.g., is part of mask $M_1$) how large the expected overlap is when drawing $x$ new objects (e.g., weights from mask $M_2$).
The hyper-geometric distribution is parametrized by the number of individual weights $mn$ and the size of the ticket $\sizeTicket<mn$. For a number of successes $x<mn$, 
its probability mass function is specified by
\begin{equation}
p(x)=\frac{{ \sizeTicket \choose x}{ mn - \sizeTicket  \choose \sizeTicket-x}}{{ mn \choose \sizeTicket}}\text{  .}
\end{equation}  
In Figure~\ref{fig:overlapseedPrune}, we plot the overlap of masks across tasks (denoted as \textbf{B} in Figure~\ref{fig:setUP}). The observed overlaps (colored dots) match the baseline (gray curve) closely, confirming our choice. Hence, we depict this baseline in all of the following plots that show overlaps between tickets.

\section{Experiments}
Previous work mentions that there might be several, not one, winning ticket in each network~\citep{anonymous2020the}.
We confirm in our setting that there are several wining tickets:
for each initialization, we run the training-pruning-resetting procedure five times, and measure the distance between the obtained masks.
Indeed, the masks are not equivalent, but vary greatly.
As they show no overlap beyond chance, we investigate whether there are shared/unused weights beyond chance, or rank correlations between masks.
We then check whether the tickets are equivalent under weight-space symmetry.
As a final sanity check, we rerun some of the experiments with partially fixed randomness and confirm that these tickets do show overlap beyond the expected baseline.

\subsection{Are winning tickets unique?}
We compare the overlaps between all tickets generated from one initialization (case \textbf{A} from Figure~\ref{fig:setUP}) after iterative pruning.
More concretely, we consider pruning levels of $80$\% (Fashion MNIST), $95$\% (CIFAR) and $98$\% (ResNet, CIFAR). 
Five training runs are compared among each other, where
we count any similarity once, yielding $10$ values for each seed.
Each experiment contains five seeds or initial weights, yielding $50$ similarities in total.
We compare the overlaps to the hyper-geometric baseline, which we plot in gray. 
To ease comparability across layers, all overlaps are shown in percent of the mask size in Figure~\ref{fig:overlapLaterPrune}.
The same plots for $50$\% sized tickets are in Appendix~\ref{sec:overlap50}.

\textbf{Results.} On Fashion MNIST (Figure~\ref{fig:overlapLaterPruneM}), the expected overlap lies around $10$\%.
The first layer's overlaps are less than $20$\%, with some overlaps at $26$\%.
The second layer's overlaps range between $5$\% and $15$\%.
The third layer's overlaps are scattered with low variance around $10$\%.
Analogously, the fourth layer's overlaps show little variance, and are scattered around $10$\%.
The last layer's overlaps vary between $3$\% and $20$\%. 

\begin{figure*}[tb]
  \begin{subfigure}[b]{0.3\textwidth}
    \includegraphics[width=\textwidth]{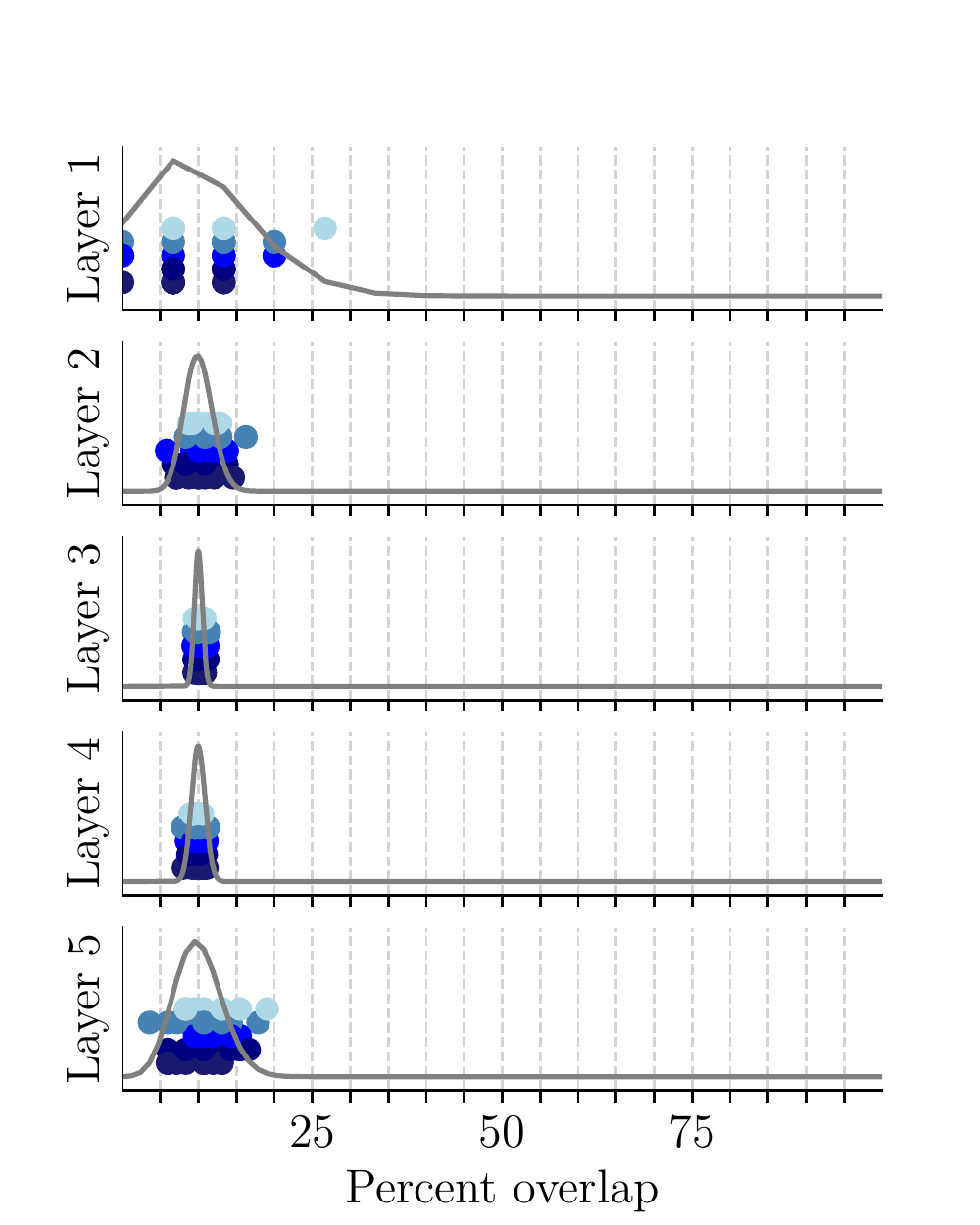}
    \caption{Fashion MNIST, $80$\% of weights pruned in $3$ iterations.}
    \label{fig:overlapLaterPruneM}
  \end{subfigure}
    \hspace{7pt}
  \begin{subfigure}[b]{0.3\textwidth}
    \includegraphics[width=\textwidth]{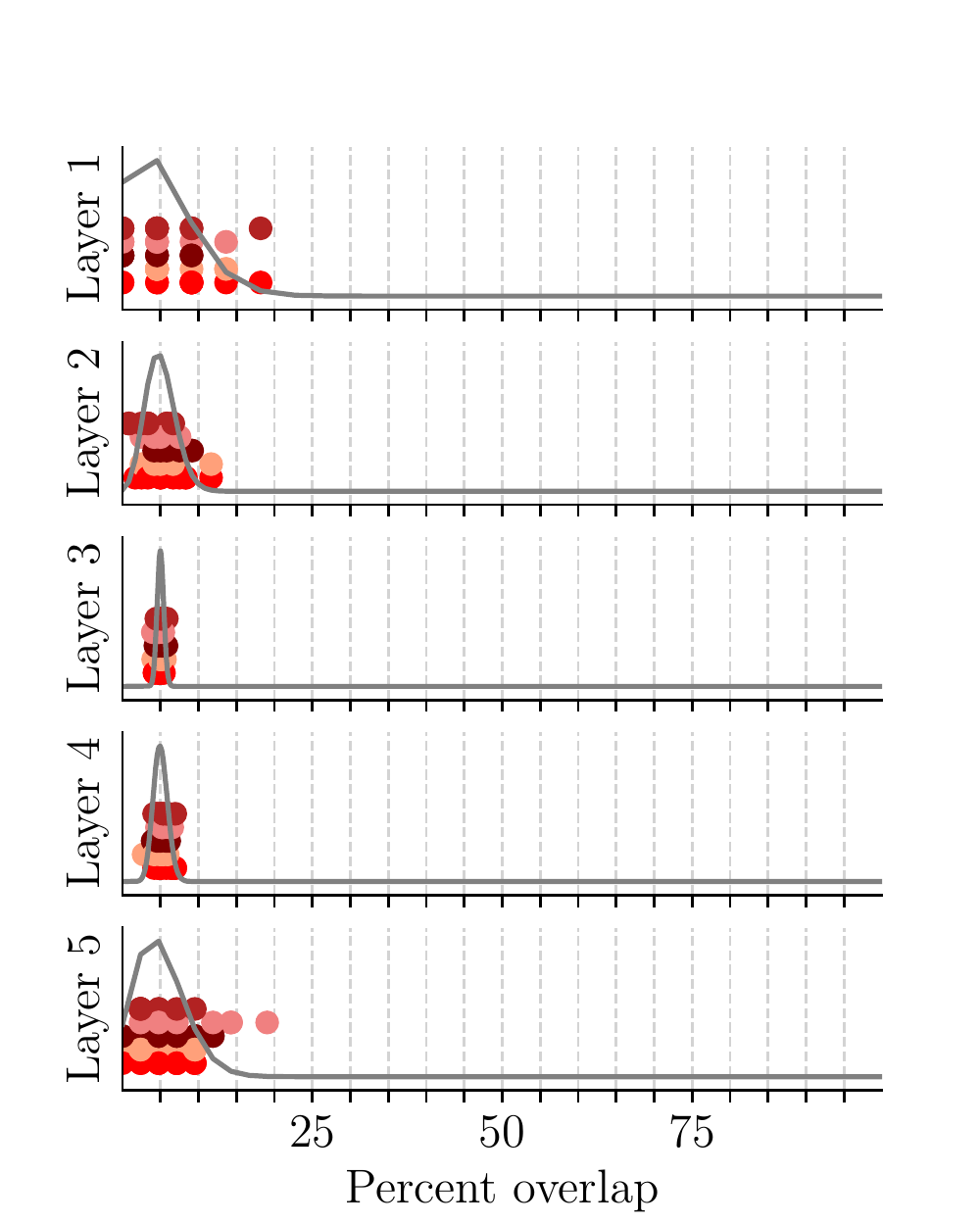}
    \caption{CIFAR10, $90$\% of weights pruned in $4$ iterations.}
    \label{fig:overlapLaterPruneC}
  \end{subfigure} 
      \hspace{7pt}
    \begin{subfigure}[b]{0.3\textwidth}
    \includegraphics[width=\textwidth]{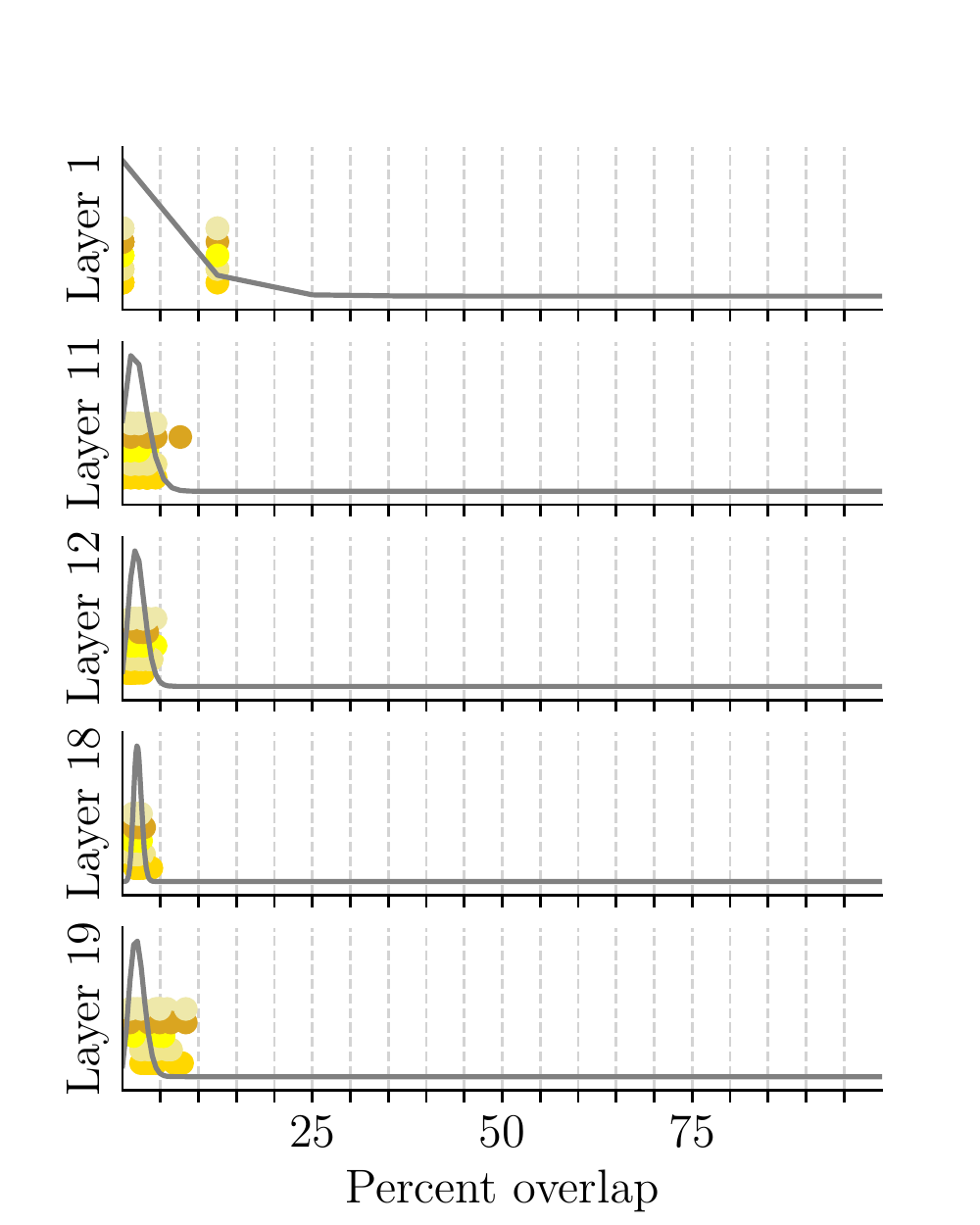}
    \caption{CIFAR10, $90$\% of weights pruned in $3$ iterations.}
    \label{fig:overlapLaterPruneR}
  \end{subfigure} 
  \caption{Percentage of overlap between pruned masks (pruning after 15 epochs) of 5 runs, each using 5 initializations. Each initialization is one shade, $y$ position is based on seed and carries no further meaning. The gray curve is the expected overlap computed by the hyper-geometric.}
  \label{fig:overlapLaterPrune}
\end{figure*}

On CIFAR (Figure~\ref{fig:overlapLaterPruneC}), we investigate a smaller ticket, sized $10$\% of the original weights.
The expected overlaps vary around $5$\%.
In particular the first layer exhibits overlaps smaller than $20$\%.
The second layer's overlaps also match the baseline, varying between $0$\% and $12$\%.
The overlaps of the third and fourth layer are scattered closely around $5$\%, where the fourth layer shows (as expected) slightly higher variance.
In the last layer, overlaps lie between $0$\% and $15$\%, with 
an outlier at $19$\%.

The expected overlaps of ResNet are around $2.5$\%, as we consider a small ticket of only $10$\% of the original weights. 
The first layer exhibits, as expected, overlaps between $0$\% and $12.5$\%.
The eleventh, twelfth and eighteenth layers overlaps vary between $0$\% and $5$\%.
In the second layer, some outliers show overlaps of $7$\%.
The last layer shows slightly higher overlaps, as expected, of up to $10$\%.

\textbf{Conclusion.} 
Tickets for the same initialization are not equivalent, and show barely more overlap than expected when we consider the hyper-geometric distribution as a baseline.

\subsection{Are there similarities beyond overlap between tickets?}
To verify that there are not structural similarities that we missed in the previous experiments, we investigate how much individual tickets for one initialization vary, and how unique they are.
In this step, we again compare within one initialization (\textbf{A} in Figure~\ref{fig:setUP}).
To compare uniqueness,
we compute for each initialization how many weights are contained in all five masks obtained from randomized training. 
Further, we investigate the inverse question: how many weights are always pruned, and never form part of a mask.
To capture another form of relationship, we plot the rank correlation between initial and final weights.
We then draw an overall conclusion.

\subsubsection{How many weights are shared?}\label{sec:initweightsA}
\begin{figure}[t]
 \begin{subfigure}[b]{0.5\textwidth}
  \includegraphics[width=\linewidth]{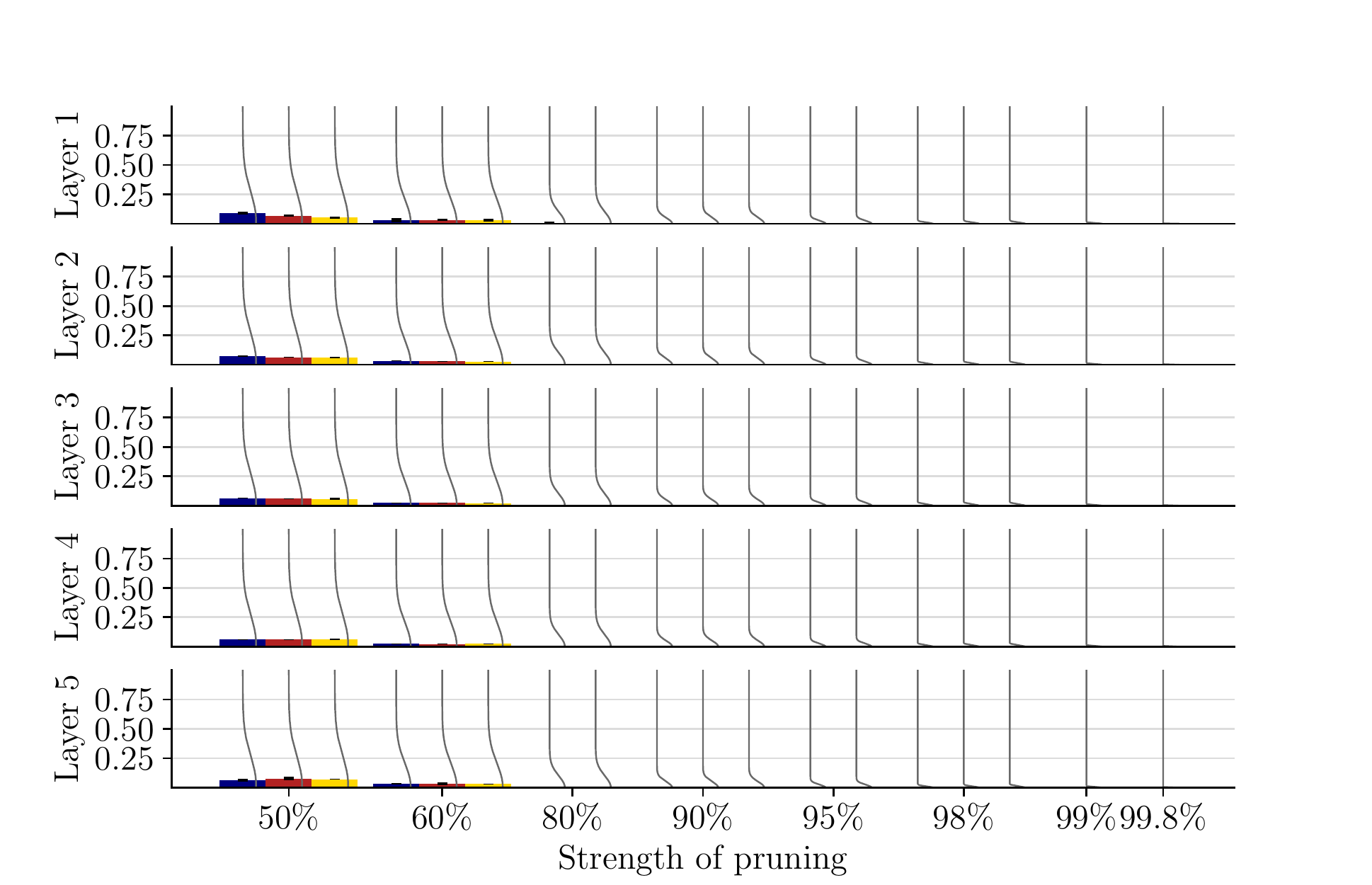}
  \caption{Percent of weights in all masks.}\label{fig:sharedWeights}
  \end{subfigure}
        \hspace{7pt}
  \begin{subfigure}[b]{0.5\textwidth}
     \includegraphics[width=\linewidth]{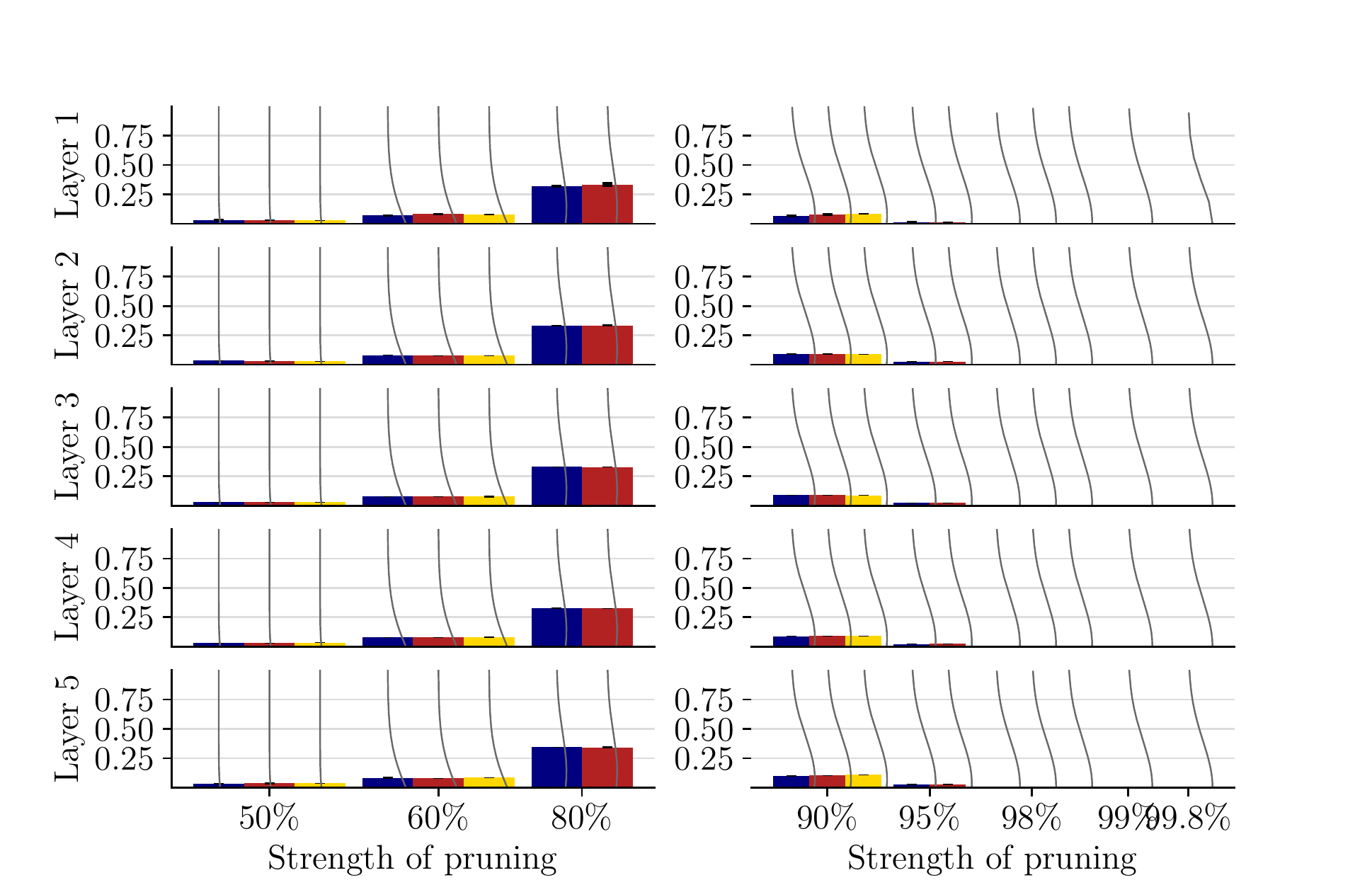}
     \caption{Percent of weights never in any mask.}\label{fig:disjointWeights}
       \end{subfigure}
       
     \caption{Influence of the initial weights on the tickets. Blue is Fashion MNIST, red CIFAR, yellow ResNet on CIFAR, and black denotes the standard deviation between the five runs.
  Gray denotes an approximated baseline. As before, the plotted layers for ResNet are first, 11th, 12th, 18th, and 19th. (a) Percentage of weights shared by all five obtained tickets (for one initialization, normalized by mask size).      
     (b) Percentage of weights not contained in neither five tickets (for one initialization). 
  Right: The left half of the plot is normalized by layer size. 
  The right half is normalized by the maximal disjoint coverage of the masks (e.g., for pruning level $98$: $2 \times 5 = 10$\%).}
\end{figure}

We first plot the weights forming part of all five masks for one initialization.
To ease understanding, we normalize the number of weights and plot percentages in Figure~\ref{fig:sharedWeights}. 
$100$\% denote the maximal number of weights that can be shared, e.g. $50$\% of the weights for pruning step one, $40$\% for pruning step two, etc.
Since we compare repeated trials (overlap between several masks),
 the hyper-geometric is not a valid baseline.
Hence we use an approximation (see Appendix~\ref{sec:baseline} for details).

\textbf{Results.} All networks generally share roughly the same, low amount of weights.  
The amount of shared weights decreases with higher pruning rate, as predicted by our baseline. 
The shared weights also slightly decrease as we go deeper into the network and
consider later layers. 
The standard deviation between different runs is generally low.
 
 The Fashion MNIST tickets exhibit the highest overlap ($10$\% shared weights) in the first layer at lowest pruning level ($50$\%).
 Both inner and later layers show slightly lower percentages of shared weights, roughly around $7$\%.
 Analogously, the number of shared weights decreases  to $0$ in the third pruning iteration.
The standard deviation between the different runs is overall very small.
 
 The shared weights for CIFAR10 are analogous, and decreases for the first layer as we iterate pruning. 
 Here as well, $0$\% shared weights are reached at pruning iteration three. 
 The initial overlap is slightly lower than Fashion MNIST, and lies around $8$\%. 
Going though the network, in this case, seems not to affect the amount of shared weights.
At the second pruning iteration, the overlaps are very low.
As for Fashion MNIST, the variance of the shared weights across runs is small.

On ResNet, the results are similar to the two smaller networks.
The shared weights do not change for the first pruning iteration when going deeper into the network.
The first layer reaches zero in pruning iteration three.
For deeper layers, the are no shared weights already in iteration two (layer 18, 19).

\subsubsection{How many weights are left out?}\label{sec:initweightsB}
We now investigate which weights never form part of any ticket.
For smaller mask sizes (pruning level $>80$\%), it is impossible that the five masks cover all weights. 
We thus normalize by the maximal coverage of all five masks or layer size, whichever is smaller.
The results are plotted in Figure~\ref{fig:disjointWeights}.
We approximate a distribution to obtain a baseline, as in the previous case (for details see Appendix~\ref{sec:baseline}).

\textbf{Results.} The amount of weights contained in no ticket is the same for Fashion MNIST, CIFAR10 and ResNet. All values lie within their corresponding expected baselines.
In the first pruning step, the amount of not-contained weights is similarly low for all cases, and lies around $2-3$\%. 
The amount increases over $10$\% (pruning $60$\%) to $30$\% when pruning $80$\% of the weights. For smaller tickets, when seen in relation to the area possibly covered, the percentages decrease again.
There are no differences for any network when considering inner or later layers. 

\subsubsection{How much do the initial weights impact winning tickets?}\label{sec:initweightsC}

\begin{figure*}[tb]
  \begin{subfigure}[b]{0.3\textwidth}
    \includegraphics[width=\textwidth]{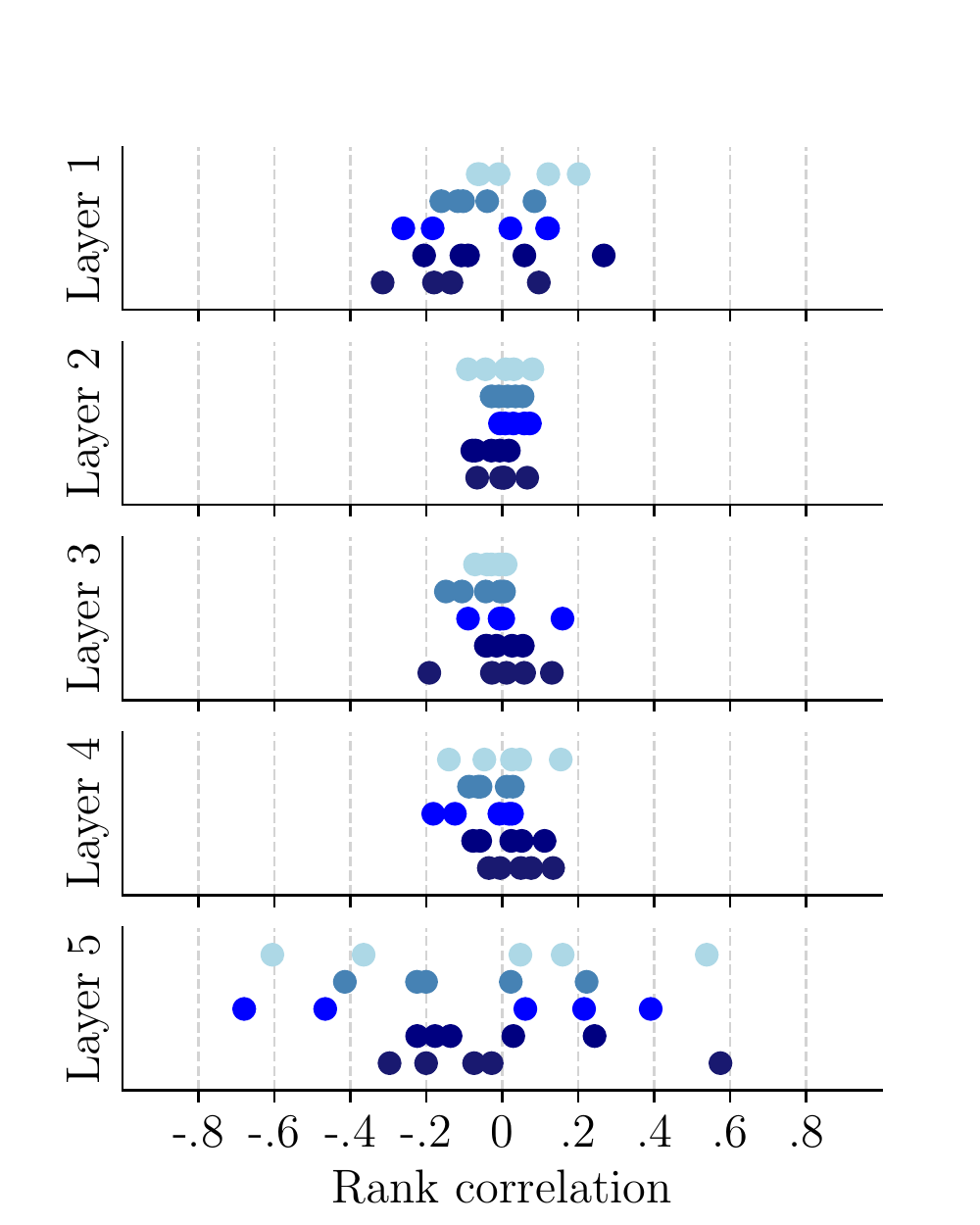}
    \caption{Fashion MNIST, $80$\% of weights pruned in $3$ iteration.}
    \label{fig:corrsLaterPruneM}
  \end{subfigure}
    \hspace{7pt}
  \begin{subfigure}[b]{0.3\textwidth}
    \includegraphics[width=\textwidth]{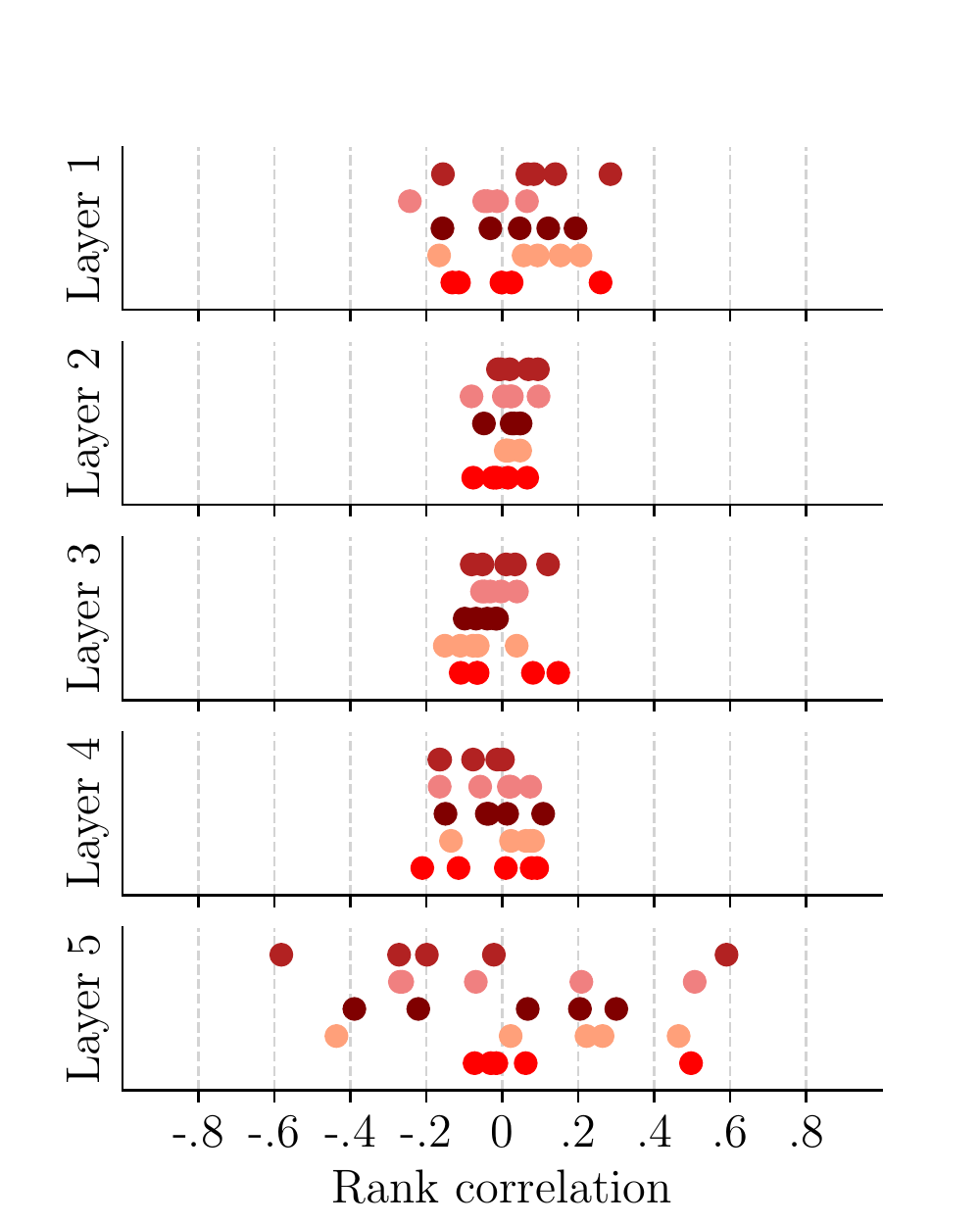}
    \caption{CIFAR10, $90$\% of weights pruned in $4$ iterations.}
    \label{fig:corrsLaterPruneC}
  \end{subfigure} 
      \hspace{7pt}
    \begin{subfigure}[b]{0.3\textwidth}
    \includegraphics[width=\textwidth]{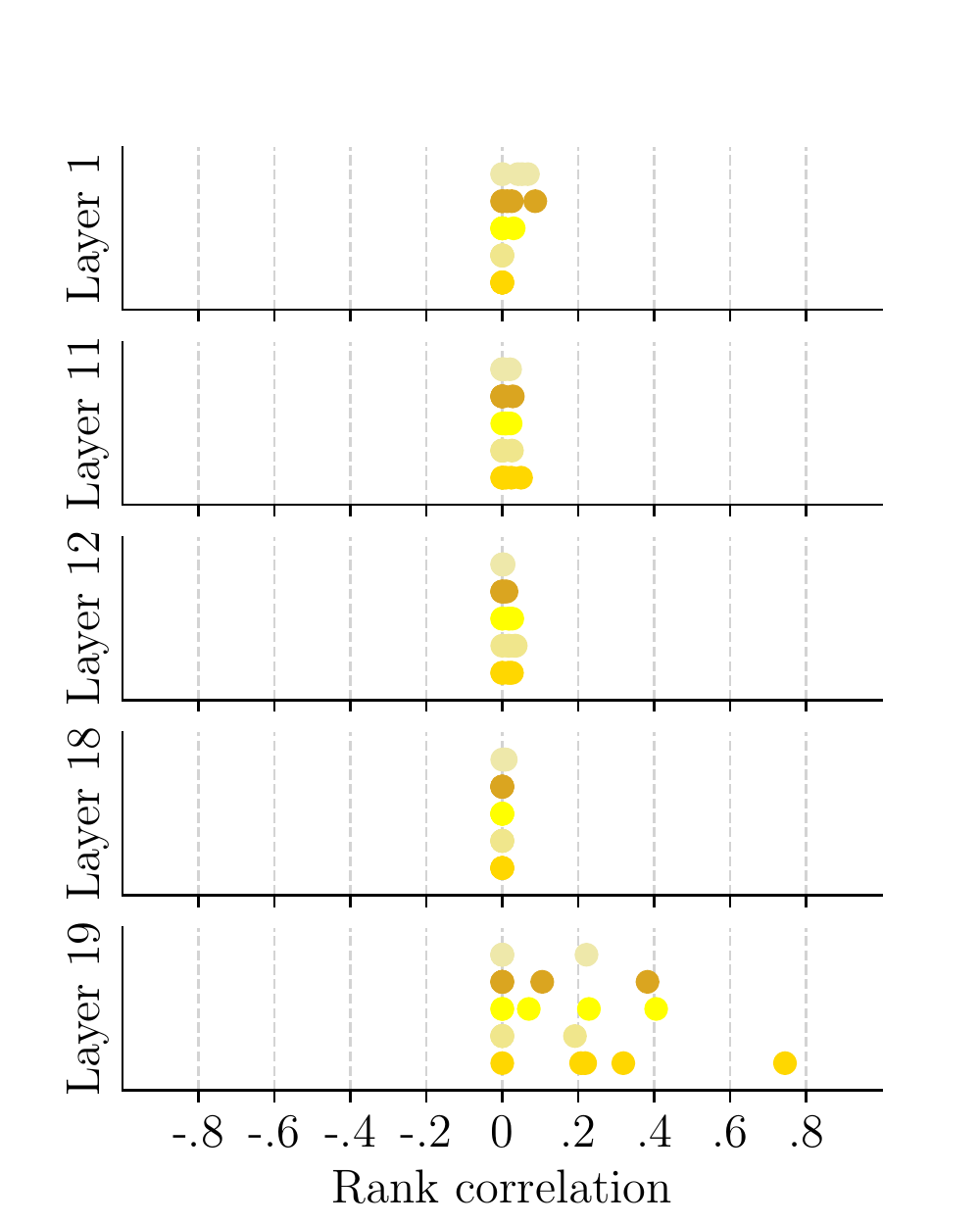}
    \caption{CIFAR10, $90$\% of weights pruned in $3$ iterations.}
    \label{fig:corrsLaterPruneR}
  \end{subfigure} 
  \caption{Rank correlation between initial weights and weights in masks (pruning after 15 epochs), each using 5 initializations. Each initialization is one shade, $y$ position is based on seed and carries no further meaning.}
  \label{fig:corrsLaterPrune}
\end{figure*}
To investigate another form of potential structure across tickets, we compute the rank correlation between the initial weights and the weights of the resulting winning tickets (setting \textbf{A} in Figure~\ref{fig:setUP}).
A high correlation of $1.0$ implies that the order is preserved, $-1.0$ means the order is inverted, $0$ that there is no relationship in terms of rank correlation. We plot the results in Figure~\ref{fig:corrsLaterPrune}.

\textbf{Results.} The Fashion MNIST correlations are centered for all layers around $0$, with differing variances. 
The first layer's correlations lie between $-0.4$ and $0.2$.
The correlations in the second layer range around $-0.05$ and $0.05$. 
The third and fourth layer exhibit correlations between $-0.2$ and $0.2$.
The last layers correlations vary greatly between $-0.7$ and $0.6$.

The small network on CIFAR ten shows, albeit for a smaller ticket, the same pattern as the Fashion MNIST network. The first layer varies between $-0.3$ and $0.4$.
The second, third, and fourth layer show little variance, with correlations scattered between $-0.2$ and $0.2$. 
The last layer show larger variation in correlations, which exhibit values between $-0.6$ and $0.6$.

In contrast to the previous two small networks, the ResNet shows almost no variance in the correlations, which are all scattered with very low variance around $0$. An exception are the first, and in particular the last layer.
The first layer shows slight positive correlations which are smaller than $0.1$.
The eleventh, twelfth and eighteenth layer exhibit correlations around zero with low spread.
The last layer's correlations, however, lie between $0$ and $0.4$, with an outlier at $0.75$.

\subsubsection{Conclusion} 
There is no significant amount of shared weights, or weights that never form part of any mask. The rank correlations between initial weights and final weights for small tickets vary and are centered around $0$. First and last layer of the small networks show larger variance in correlations. The ResNet shows less variance of correlations, with the exception of the first and the last layer. In general for ResNet, however, correlations are zero or positive, never negative. 

\subsection{Are tickets in fact variations over the same network?}\label{sec:weightSpaceSymmetry}

\begin{wrapfigure}{l}{0.62\textwidth}
  \vspace{-15pt}
    \includegraphics[width=0.6\textwidth , trim={0 7.74cm 0.4cm 1.3cm},clip]{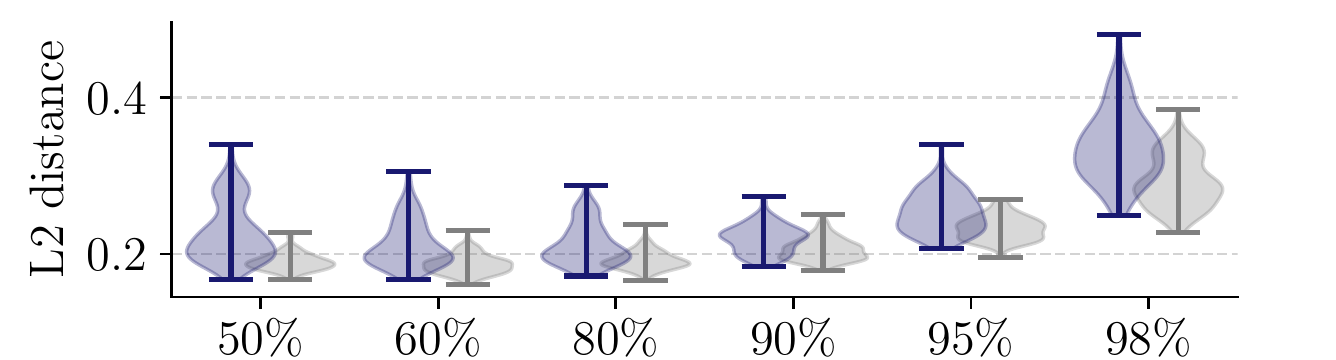}
  \caption{CKA similarity between winning tickets within (blue, left) and across (gray, right) seeds for Fashion MNIST.}
  \label{fig:CKAMain}
    \vspace{-13pt}
\end{wrapfigure}
One might be tempted to explain the differences in tickets by weight-space symmetry: The differing networks would then, in fact, be just variants of the same network. To show that this is not the case, we take advantage that networks, if equivalent in weight-space, will yield the same output (as the variations do not touch functionality). We computed the $L_2$ distance and depict it on the Fashion MNIST network, where outputs are most similar. We plot the distances of outputs of tickets in for one seed (blue, \textbf{A} in Figure~\ref{fig:setUP}) as well as distance among seeds (gray, \textbf{B} in Figure~\ref{fig:setUP}) in Figure~\ref{fig:CKAMain}. The full results, including the small network on CIFAR and the ResNet as well as the CKA measure by \cite{kornblith2019similarity} can be found in Appendix~\ref{sec:ckaFull}.

\textbf{Results.} There are not two tickets yielding the exact same input.The distances between tickets first decreases as the pruning level increases, then decreases, somewhat similar to the accuracy. The distances across seeds remain stable for the first three pruning iterations, and then increase. The distances across seeds are overall lower than within seeds.

\textbf{Conclusion.} The distance between the outputs increases as we prune iteratively and harvest smaller tickets. As the distance is never zero, we can refute the hypothesis that different tickets are instances of the same network under the weight-space hypothesis.

\subsection{What is the effect of constraining randomness?}\label{sec:fixed}
So far, we did not fix randomness at all. Generally, randomness is removed from experiments to increase reproducibility. In our setting, with no randomness, we expect the tickets to overlap perfectly. We are thus interested in the gray zone, where randomness is decreased, but not entirely fixed. We repeat all previous experiments, and depict a subset of the results in Figure~\ref{fig:Fix}. All results for the various settings can be found in the subsections of Appendix~\ref{sec:bonus}. 

\begin{figure*}[tb]
  \begin{subfigure}[b]{0.3\textwidth}
    \includegraphics[width=\textwidth]{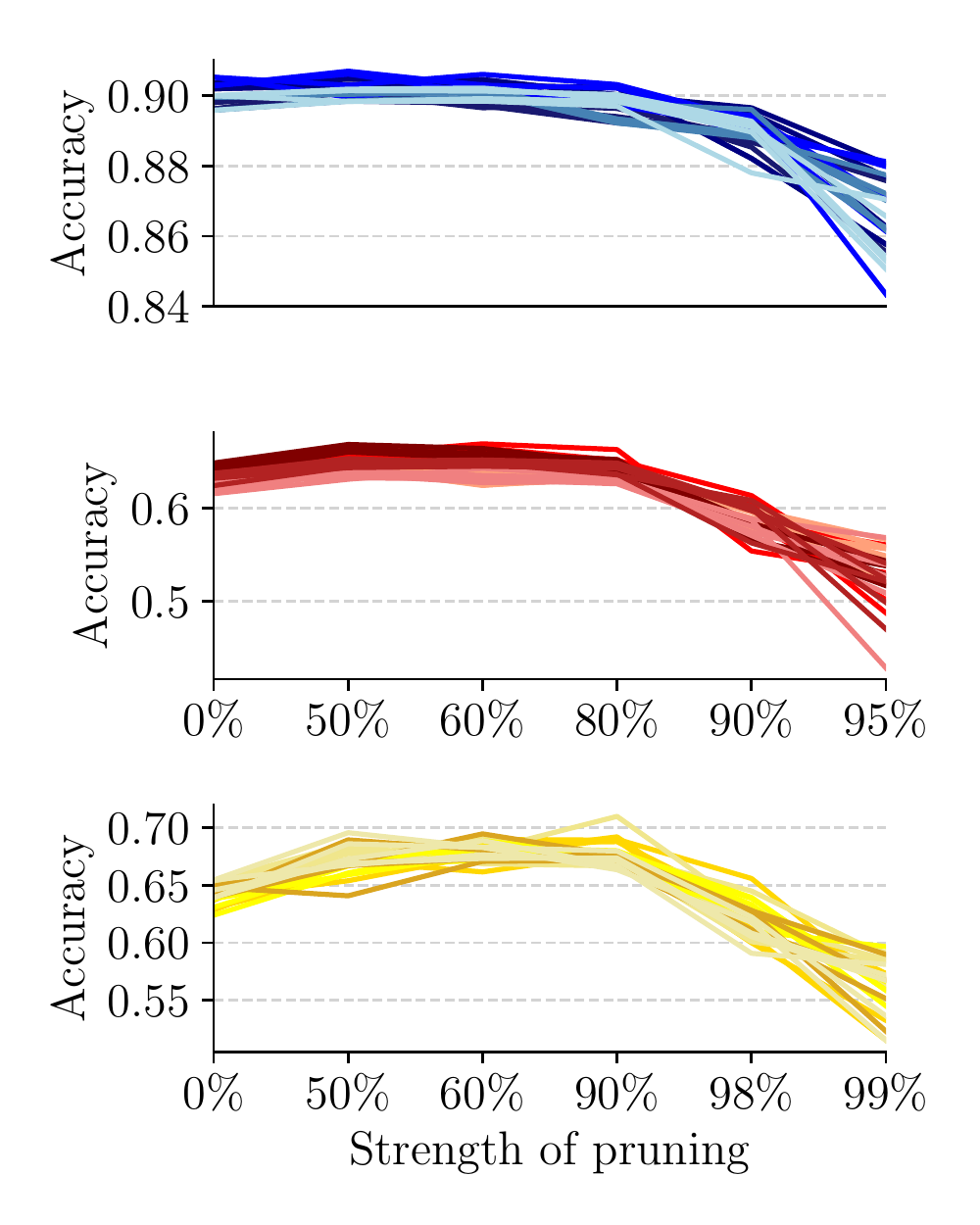}
    \caption{Accuracies.}
    \label{fig:FixAcc}
  \end{subfigure}
     \hspace{7pt}
    \begin{subfigure}[b]{0.32\textwidth}
    \includegraphics[width=\textwidth]{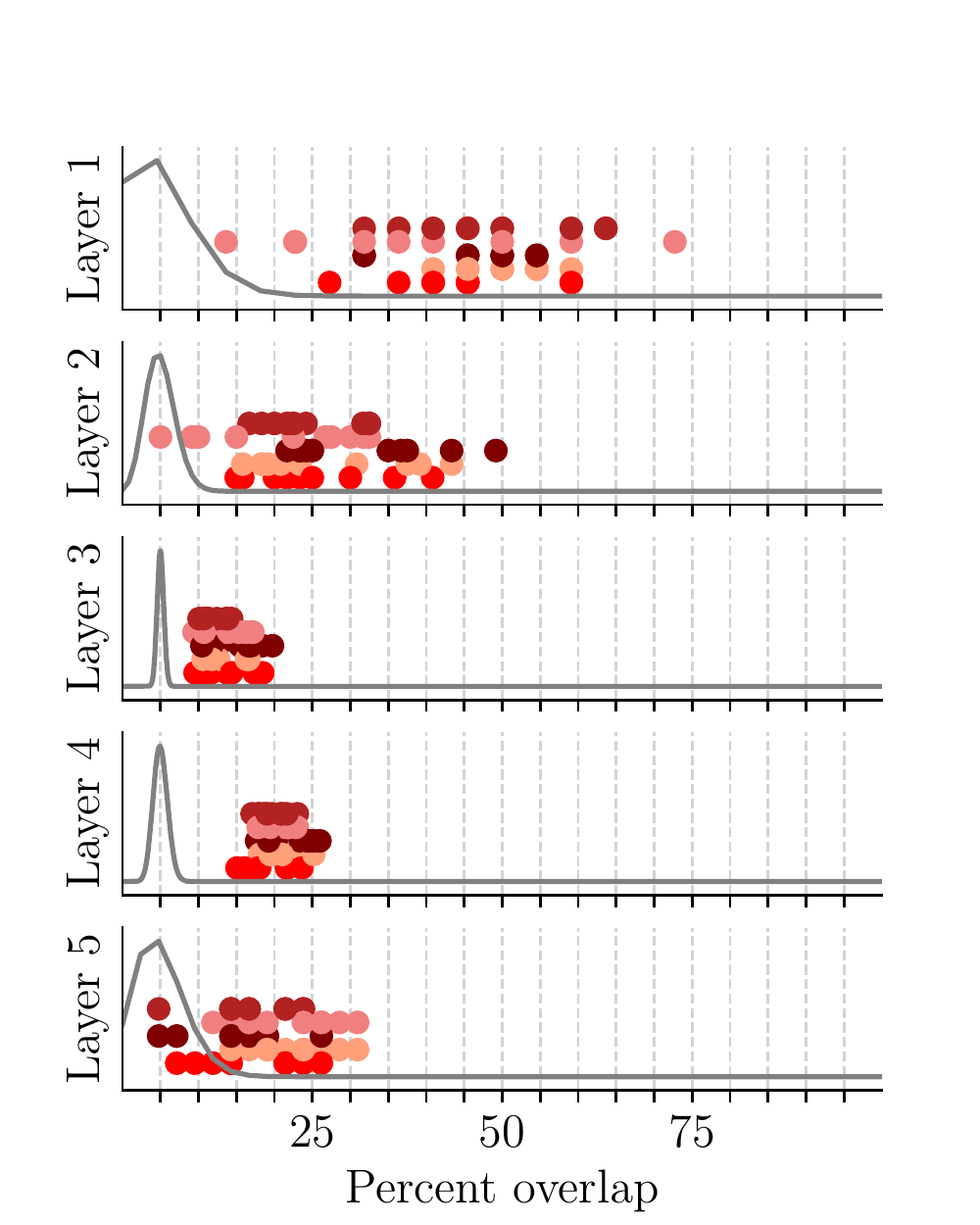}
    \caption{Overlaps.}
    \label{fig:overlapsFixC}
  \end{subfigure} 
    \hspace{7pt}
  \begin{subfigure}[b]{0.32\textwidth}
    \includegraphics[width=\textwidth]{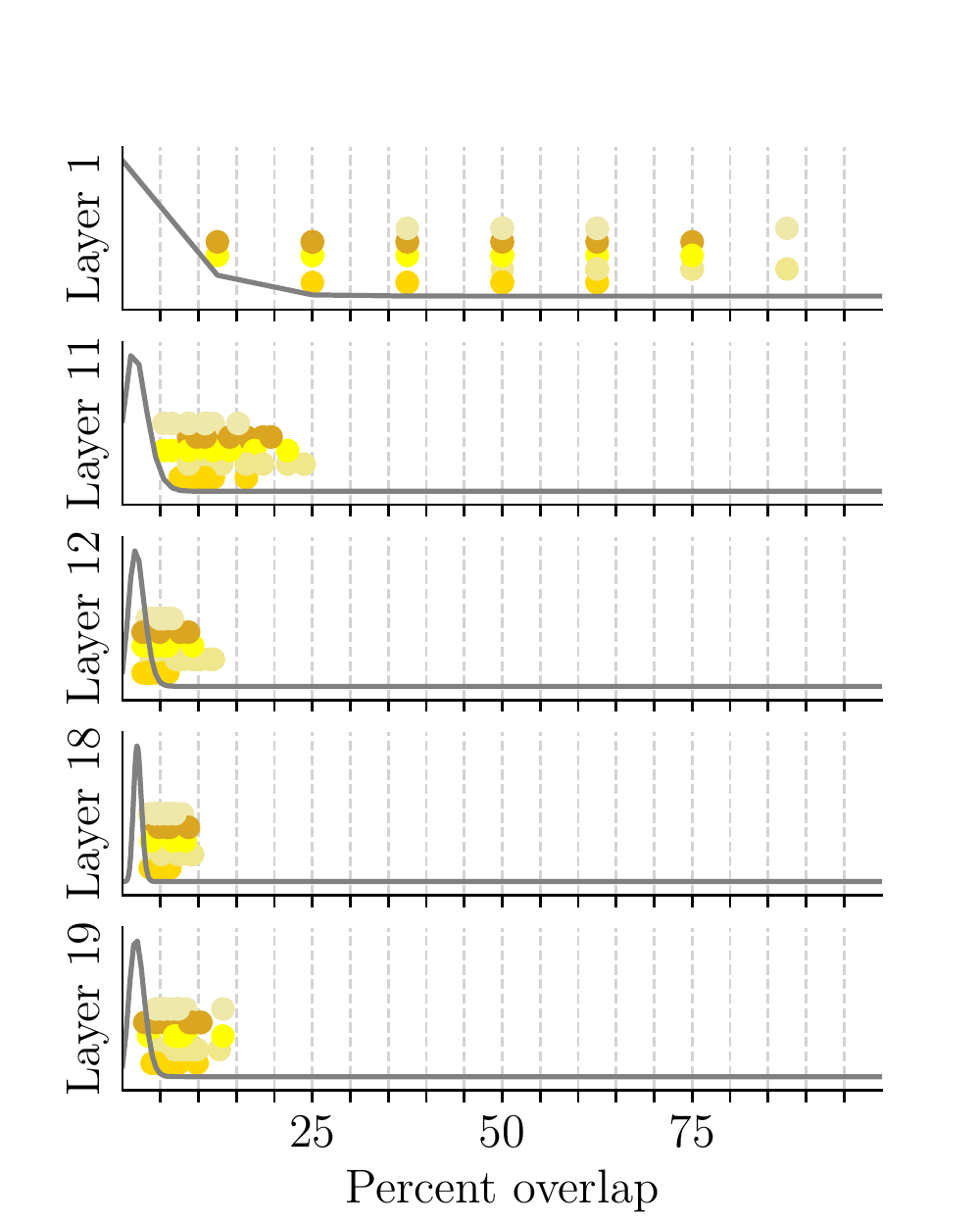}
    \caption{Overlaps.}
    \label{fig:overlapsFixR}
  \end{subfigure} 
    \begin{subfigure}[b]{0.3\textwidth}
    \includegraphics[width=\textwidth]{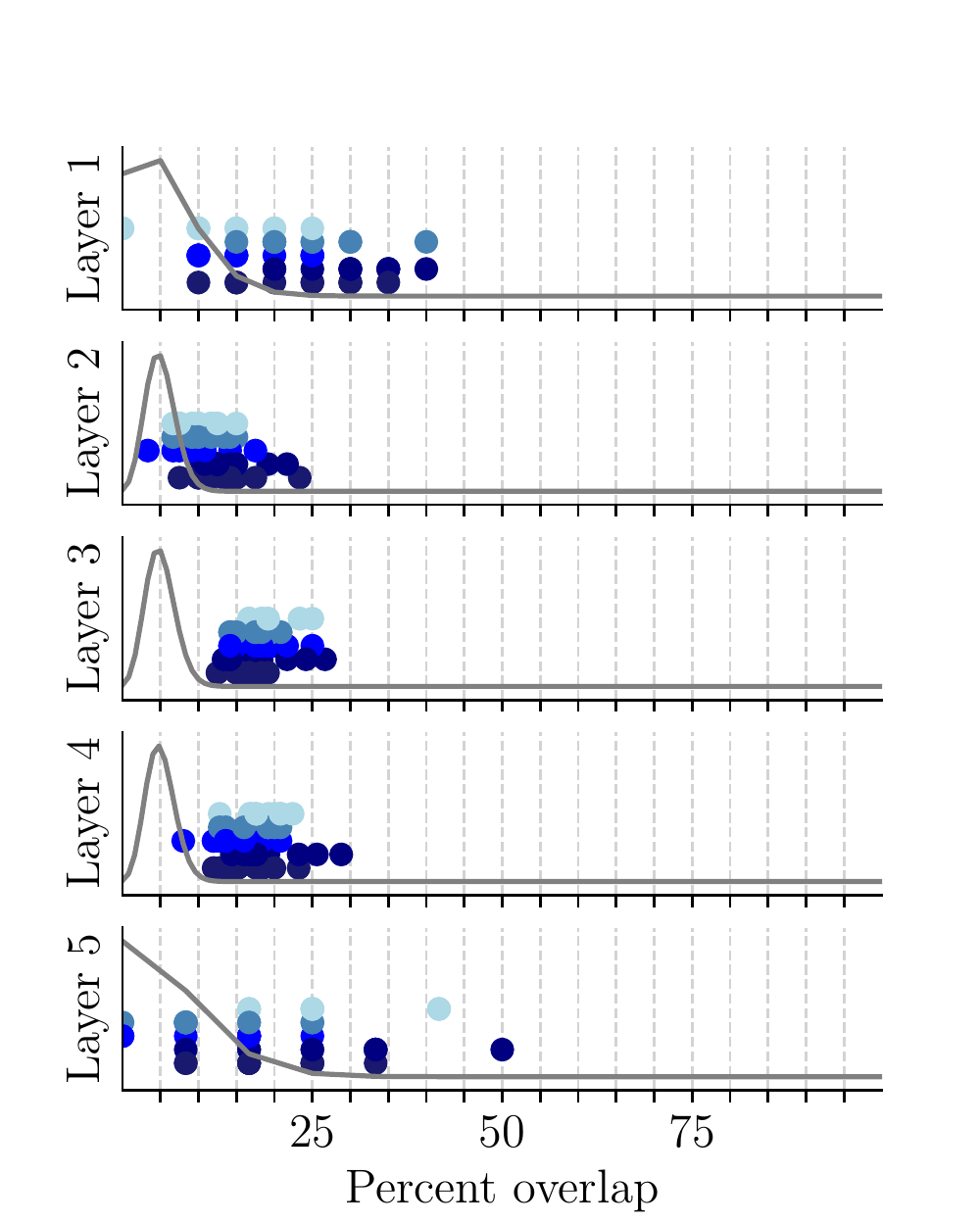}
    \caption{Overlaps.}
    \label{fig:overlapsFixFM}
  \end{subfigure}
     \hspace{7pt}
    \begin{subfigure}[b]{0.32\textwidth}
    \includegraphics[width=\textwidth]{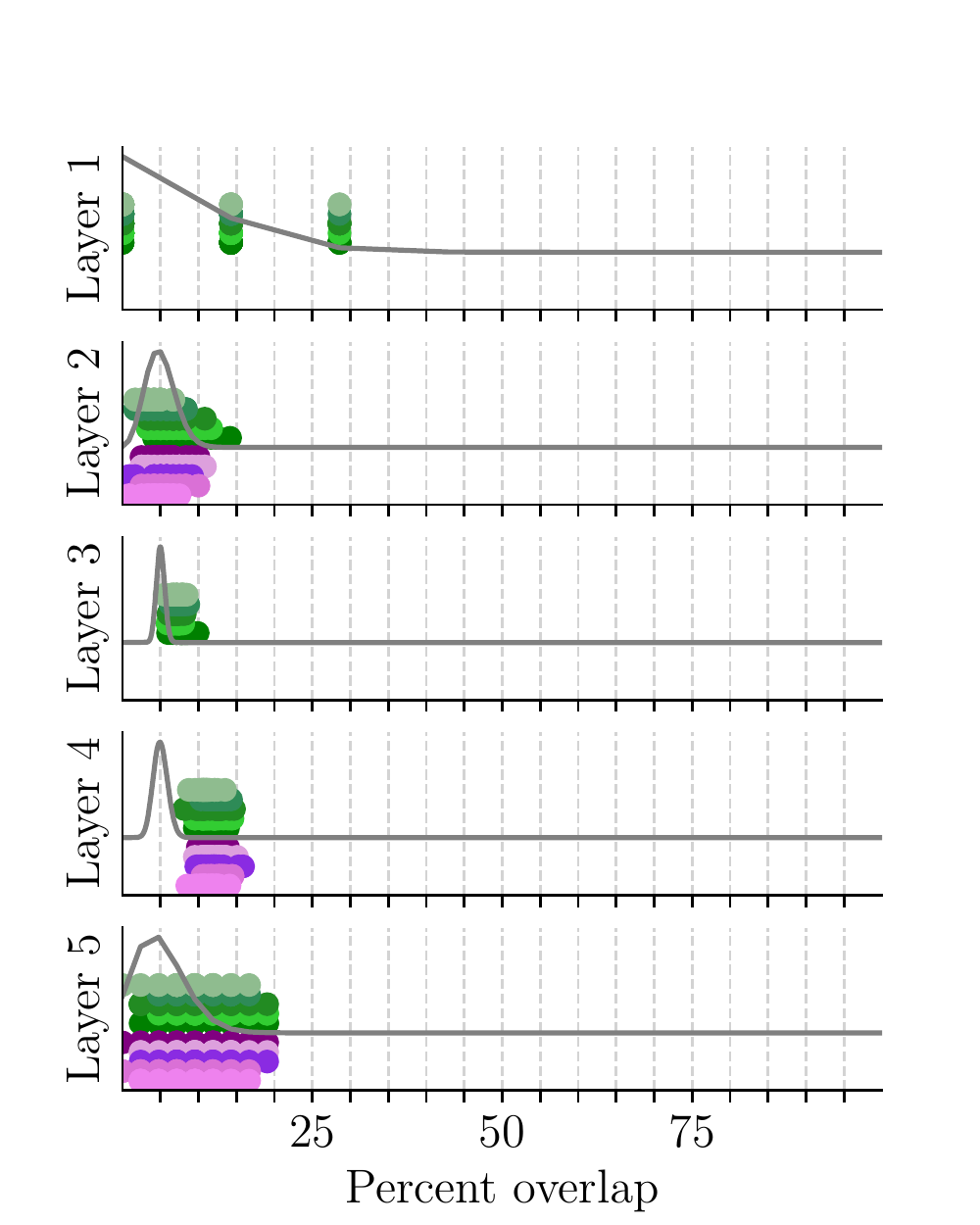}
    \caption{Overlaps.}
    \label{fig:overlapstask}
  \end{subfigure} 
    \hspace{7pt}
  \begin{subfigure}[b]{0.32\textwidth}
    \includegraphics[width=\textwidth]{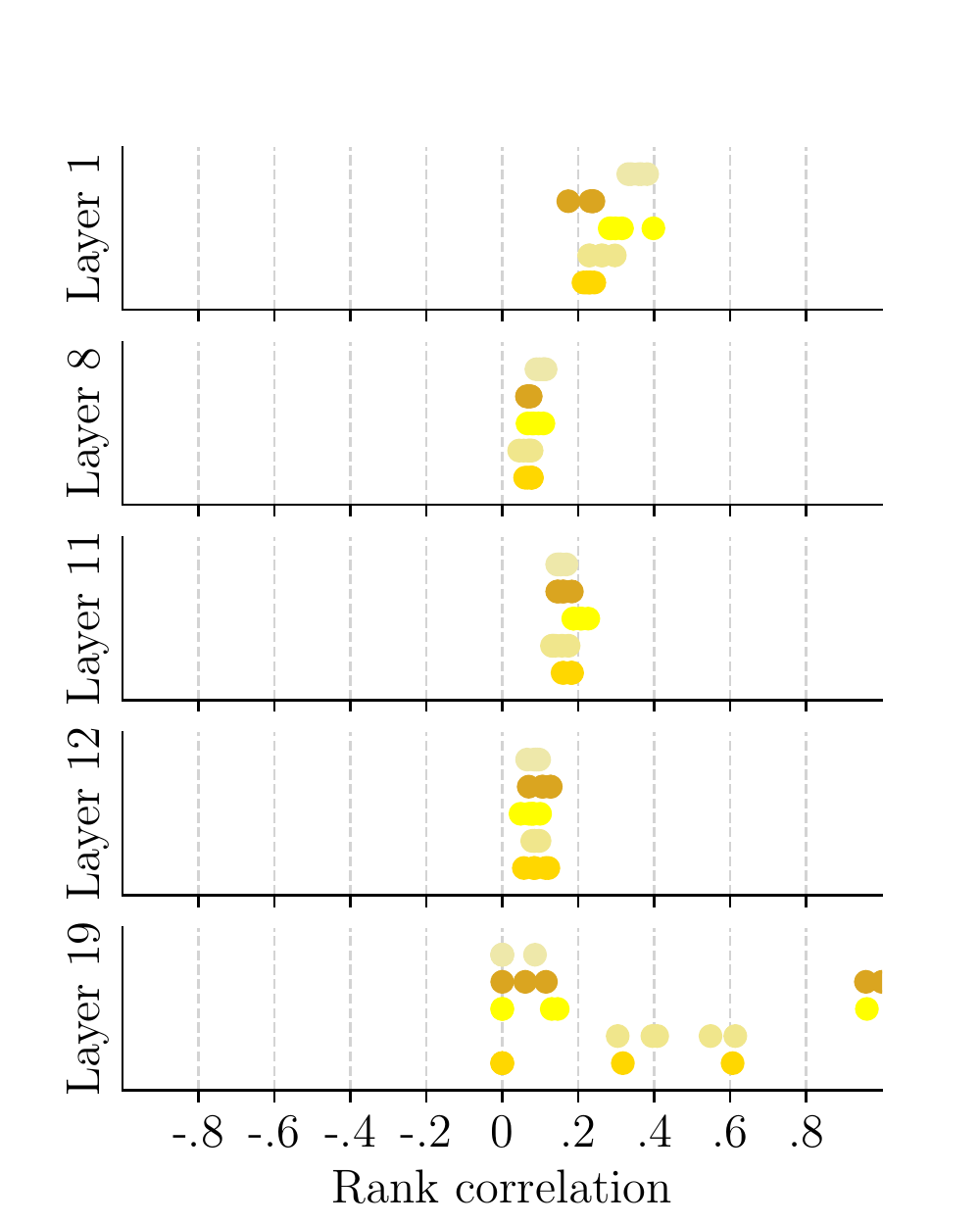}
    \caption{Rank Correlation.}
    \label{fig:corrsFix}
  \end{subfigure} 
  \caption{Repeating the previous experiments with slightly fixed randomness. 
  (a): accuracies of different pruning stages.
  (b): overlaps between CIFAR tickets in the fourth pruning iteration. 
  (c): overlaps between ResNet masks in the third pruning iteration. 
  (d): overlaps between networks on Fashion MNIST with equal-sized layers. 
  (e): overlaps across tasks with same initialization at pruning level 4: green denotes overlap between Fashion MNIST and MNIST, purple between CIFAR and Fashion MNIST.
  (b) rank correlations on a CIFAR10 ResNet at pruning iteration three.  }
  \label{fig:Fix}
\end{figure*}

\textbf{Results - Accuracies.} Figure~\ref{fig:FixAcc} shows that accuracies are overall higher, and now increase for initial pruning iterations of small networks. 
Instead of a steady incline, the accuracies increase at the first pruning step, and decrease later, after the third pruning step. This holds irrespective of task or architecture, and is similar to the ResNet without fixed randomness. 

\textbf{Results - Overlaps.} Also, as expected, the overlaps between tickets are now, due to decreased randomness, larger than expected by the hyper-geometric. We show results on   a small CIFAR network (see Figure~\ref{fig:overlapsFixC}) and a ResNet18 (see Figure~\ref{fig:overlapsFixR}). 
The overlaps vary in both cases most for the first, second/eleventh and last layer. In general, the spread of the overlaps seems related to the variance of the hyper-geometric baseline. To verify this, we depict in Figure~\ref{fig:overlapsFixFM} a network on Fashion MNIST where all inner layers have approximately the same size (e.g., $2400$ or $2500$ weights, first layer $400$, last layer $250$). Indeed, the variances of all overlaps are once again strongly correlated with the baseline. The average overlap seems not to rely on the random baseline, though.

\textbf{Results - Dependence on data.} To check how strong the influence of the fixed randomness is, we compare the overlap with fixed randomness across different tasks. As the small CIFAR network, due to larger input dimensionality, can only be compared in layer two, four and five, we train additional networks on the MNIST dataset~\cite{lecun1998gradient}. We depict the overlap between CIFAR and Fashion MNIST networks in purple, and MNIST and Fashion MNIST in green.
The expected overlap is around $5$\%.
For the first layer, the overlap is as expected. 
The second layer exhibits marginally higher overlaps.
The third layer shows slightly higher overlaps, ranging between $5$\% and $10$\%.
Layer four's overlaps range between $7.5$\% and $15$\%, and are clearly distinct from the expected values.
The last layer shows again only slightly higher overlap than expected, ranging between $0$\% and $20$\%.

\textbf{Results - Rank correlations}
Last but not least, the rank correlation is now not centered at or around zero. From Figure~\ref{fig:corrsFix}, we see that all rank correlations are positive.
The first layers correlations are between $0.15$ and $0.4$. 
The eighth and twelfth layers correlations vary around $0.1$. 
The eleventh layer shows slightly higher correlations which lie between $0.1$ and $0.2$.
The last layer exhibits large variability of the rank correlations, ranging between $0$ and $1.0$.

\textbf{Conclusion.} Decreasing randomness also decreases the differences between tickets and leads to more overlap between tickets, even if networks are trained on different tasks. The accuracy also benefits from decreased randomness, and is slightly higher. 

\subsection{Implications}
Having concluded the experiments, we now summarize the main findings and set them in relation.

When different initializations are used, the overlap between masks is predicted by the hyper-geometric distribution. 
Using the same initial weights without fixed randomness corresponds to the hyper-geometric as well. There are no other correlations, shared or excluded weights beyond chance. 
The networks are also not instances of the same network under weight-space symmetry.
When randomness is fixed, tickets show higher overlap, and also higher accuracy. This might explain why starting the training for the pruned network not in iteration $0$ but slightly later is successful~\citep{frankle2019lottery}.

\textbf{Beyond lottery tickets.} Overall, winning tickets are a phenomenon that arises \emph{after} having trained the network. 
These tickets specify a sub-network that alone achieves as good performance as the whole network. 
Since we can obtain completely different tickets for the same initial weights, however,
\emph{before} training, there is rather a distribution of capable sub-networks.
The winning ticket is then just one instance drawn from this distribution.
Future work has to determine if we can describe this distribution, in particular before the network is trained on the data.
This would allow to choose better sub-networks, or improve initialization, and train smaller networks from the start.
   

\section{Conclusion}
In this paper, we revisited the lottery ticket hypothesis.
There are winning sub-networks,
however, from the perspective of an untrained network, there are many potentially capable sub-networks, of which different are picked if randomness is not constrained during training.
We investigated how tickets differ, and found there are no shared or unused weights, and no rank correlation between tickets generated from one initialization.
When randomness is partially fixed, however, there are overlaps beyond the chance baseline, even when training on different tasks.
Tickets from partially fixed randomness during training outperform their counterparts on fully randomized training and pruning.
We hope that future work will uncover more properties the distribution of capable sub-network in the initial weights, and deepen our understanding of what winning tickets are and how they emerge. 

%
%

\section*{Acknowledgments}
The authors would like to thank the anonymous reviewers for their helpful feedback.
This work was supported by the German Federal Ministry of Education and
Research (BMBF) through funding for the Center for IT-Security,
Privacy and Accountability (CISPA) (FKZ: 16KIS0753).

\bibliographystyle{unsrtnat}
\bibliography{lit}

\clearpage

\appendix

\section{Uniqueness of Larger Tickets}\label{sec:overlap50}

\begin{figure*}[!tb]
  \begin{subfigure}[b]{0.3\textwidth}
    \includegraphics[width=\textwidth]{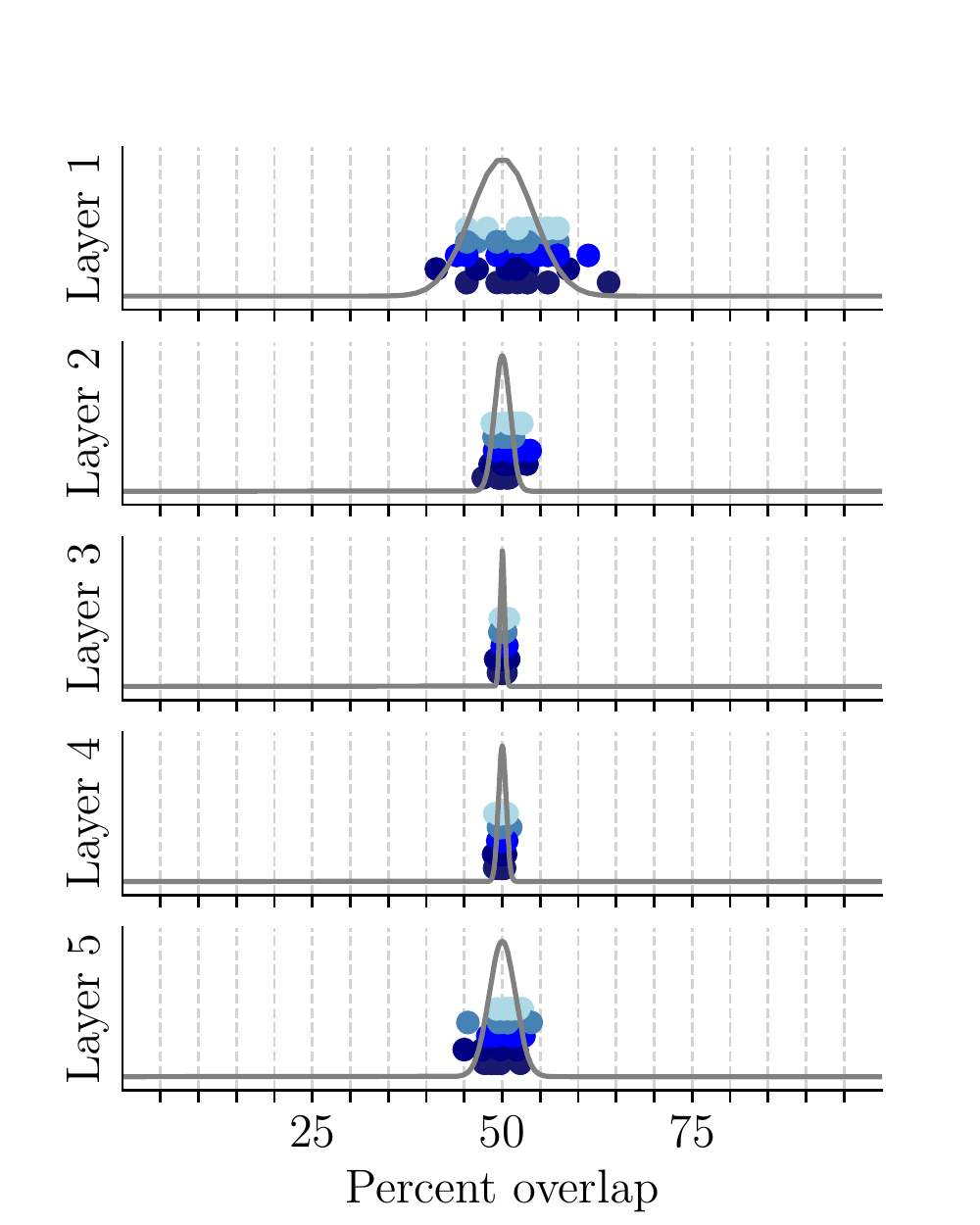}
    \caption{Fashion MNIST}
      \label{fig:overlapFirstPruneM}
  \end{subfigure}
        \hspace{7pt}
  \begin{subfigure}[b]{0.3\textwidth}
    \includegraphics[width=\textwidth]{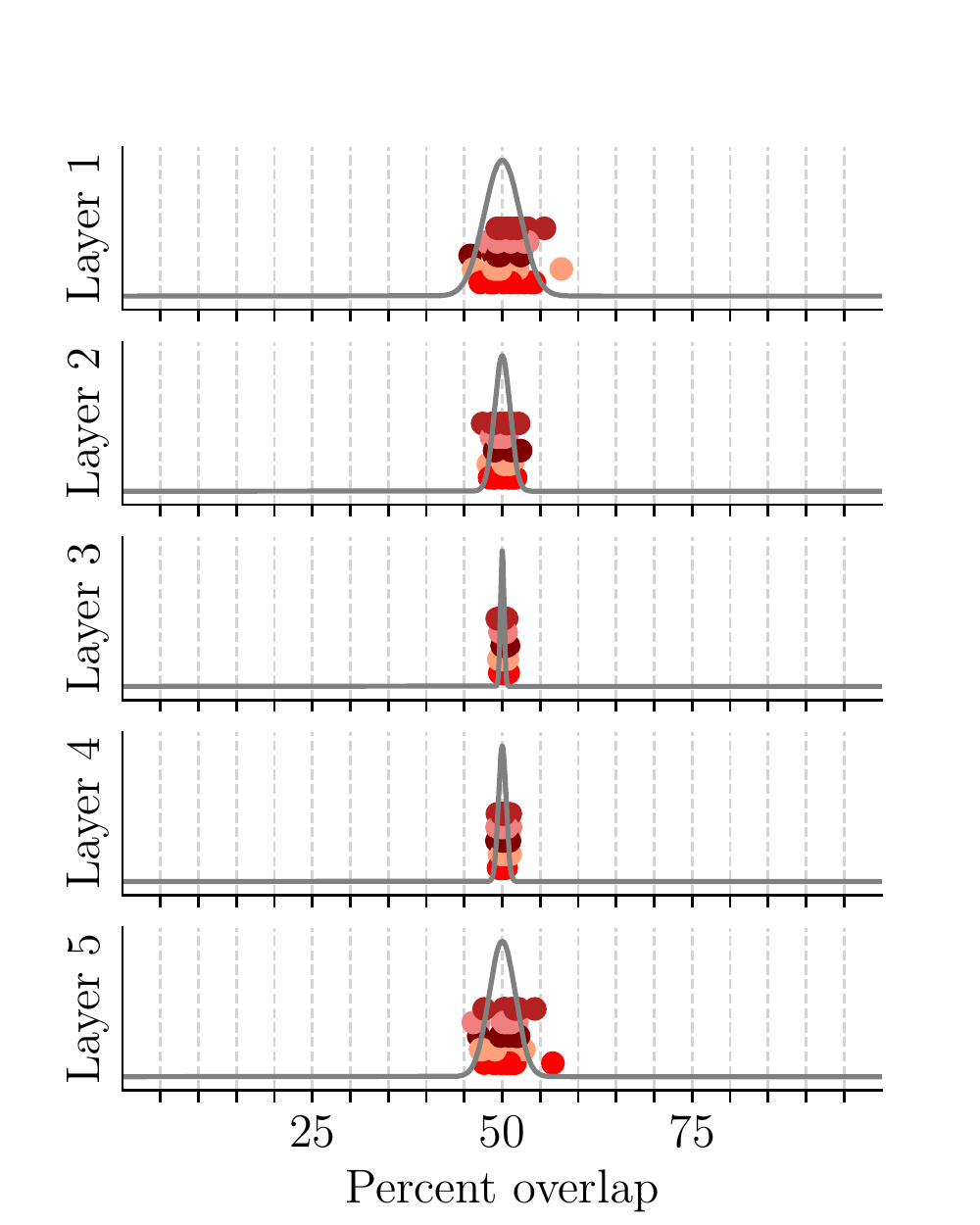}
    \caption{CIFAR10}
      \label{fig:overlapFirstPruneC}
  \end{subfigure} 
        \hspace{7pt}
    \begin{subfigure}[b]{0.3\textwidth}
    \includegraphics[width=\textwidth]{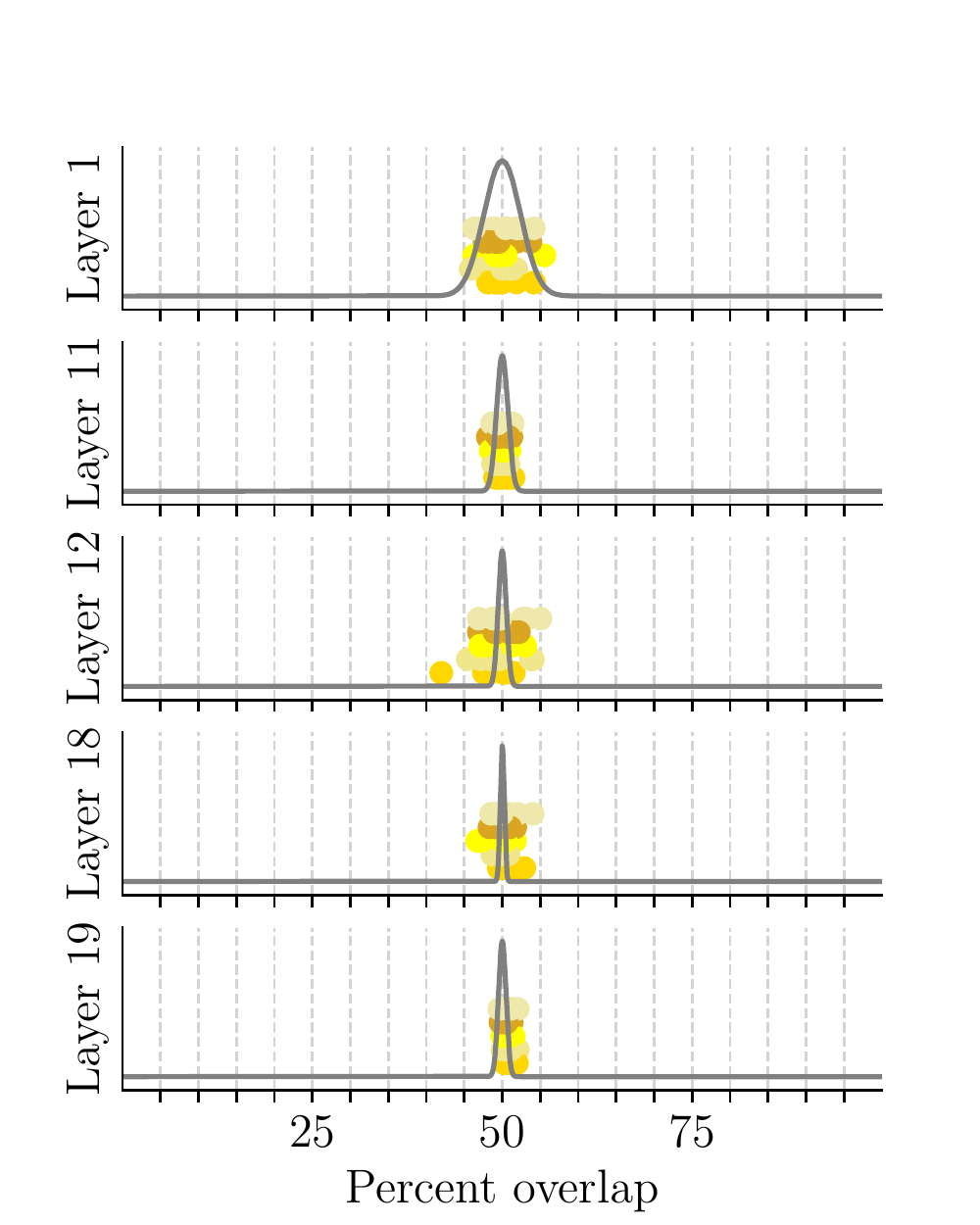}
    \caption{CIFAR10 (ResNet)}
      \label{fig:overlapFirstPruneR}
  \end{subfigure} 
  \caption{Percentage of overlap between pruned masks (when pruning $50$\% of the weights after 15 epochs) of 5 runs, each using 5 initializations. Each initialization is one shade, $y$ position is based on seed and carries no further meaning. Gray curve is the expected overlap.}
  \label{fig:overlapFirstPrune}
\end{figure*}

We compare the overlaps between all tickets generated from one initialization (case \textbf{A} from Figure~\ref{fig:setUP}) 
 at the first pruning step ($50$\% weights pruned). 
 We compare the
 similarities to the hyper geometric baseline, which we plot in gray. 
 To ease comparability across layers, 
 all overlaps are shown in percent of the mask size. We plot the results in Figure \ref{fig:overlapFirstPrune}.

\textbf{Results.} We first discuss the results on Fashion MNIST (Figure~\ref{fig:overlapFirstPruneM}). 
In general, the percentage of shared weights is very similar to the hyper-geometric baseline.
The expected overlap lies around $50$\%, where the variance depends on the number of weights in the layer.
The first layer's overlaps start around $40$\% and go up to $65$\%, which is slightly higher than the baseline.
The second layer shows overlaps between $45$\% up to $55$\%.
The third and fourth layer's similarities are scattered around $50$\%.
The last layer shows again higher variance, and varies between $45$\% and $55$\%.
 
The results are similar on CIFAR (Figure~\ref{fig:overlapFirstPruneC}). 
The overlaps of the first layer lie between $45$\% and $55$\%, with outliers between $55$\% and $60$\%.
The second layer's similarities vary between $47$\% and $53$\%.
The third and fourth layer show less variance
 and are scattered around $50$\%.
The last layer shows again slightly higher spread, with overlaps between $45$\% and $55$\% and outliers at $57$\%.

The results on ResNet (CIFAR, Figure~\ref{fig:overlapFirstPruneR}) are similar.
The overlaps of the first layer lie between $45$\% and $55$\%.
The eleventh layer show little variance, all overlaps are around $50$\%.
The twelfth layer shows larger variances ranging between $45$\% and $55$\% with an outlier at $42$\%.
The eighteenth layer is similar to the previous one, however without outliers. 
The last layer shows little variance of the overlaps, which are however slightly biased towards more overlap and are between $50$\% and $55$\%.

\textbf{Conclusion.} The overlaps between the masks are as random as predicted by the hyper-geometric, and comparable to masks drawn from different initializations.

\newcommand{\N}[0]{mn} 
\newcommand{\K}[0]{\tau} 
\newcommand{\n}[0]{|t|} 
\section{Baselines for Section~\ref{sec:initweightsA} and Section~\ref{sec:initweightsA}}\label{sec:baseline}
We now derive an approximation to compute the overlap of repeated trials.
Consider that we cannot use the binomials, or other distribution with replacement here.
The hyper-geometric from the background section, however, does not allow for repeated trials with replacement.
We hence approximate the baseline based on the hyper-geometric.
First, we define the population size (size of the weight matrix) $\N$, the size of the successes $\K$ and the number of draws $\n$.
Initially and in the main paper, $\n = \K$.
We now recap the mean and variance of the hyper-geometric given as 
\begin{equation}
\mu = \sizeTicket\frac{\sizeTicket}{mn} \quad \text{  and} \quad \sigma = \sizeTicket\frac{\sizeTicket}{mn}\frac{N-\sizeTicket}{N}\frac{N-\sizeTicket}{N-1}\text{  .}
\end{equation}  
Our main task is to compute an approximation for $1)$ the number of all drawn weights, $2)$ the number of weights that were never drawn.

\subsection{Weights shared among all tickets.} 
We compute $\K$ as only overlap among tickets. For the second drawn ticket, the expected overlap with the first is
\begin{equation}
\K_1 = \n \frac{\n}{\N} \pm \n \frac{\N-\n}{\N} \frac{\N - \n}{\N -1}\text{\space .}
\end{equation}
We focus on the average overlap first, and then derive the variance.
For the next ticket, we write the probability that it overlaps with the weights that were shared by all previous tickets. Hence, the mean for $\K_i$ with $i>1$ is
\begin{equation}\label{eq:all}
\K_i = \n \frac{\K_{i-1}}{\N}\text{\space .}
\end{equation}
Since we run $5$ independent runs, we need $K_4$ ($5-1$, as the first ticket does not overlap) in this case.

We now approximate the variance. 
After \cite{pukelsheim1994three}, $3 \times$ the variance of uni-modal distributions contains roughly $95$\% of the distribution's mass.
We hence  additionally compute the overlap with $3 \times$ the hyper-geometric's variance,
\begin{equation}
\K_i^{\max} = \n \frac{\K_{i-1}^{\max}}{\N} + 3 \n \frac{\N-\K_{i-1}^{\max}}{\N} \frac{\N - \n}{\N -1}\text{\space .}
\end{equation}
The difference between $\K_i^{\max} $ and $\K_i$ can then be used to approximate the variance of the underlying distribution.
Since we do not know the shape of the true distribution, we will use the estimated mean and variance as parameters for a normal distribution. 
As however $95$\% of the mass are contained in $2 \times$ the variance of the normal, we use $0.5 \times (\K_i^{\max} -\K_i)$ as variance to preserve the ratios.

\subsection{Weights in neither ticket.}
The hyper-geometric only determines the overlap between tickets. We want to compute the weights that are in at least one ticket.
By the complementary probability, we then obtain the number of weights that are in no ticket. 
The subsequent derivation differs only slightly from Equation~\ref{eq:all}. To obtain all weights that are covered, we add the overlaps and also remaining, not overlapping weights.
The first ticket counts fully, hence $\K_0 = \n$.
For $i>0$, we write as above
\begin{equation}
\K_i = (\n - \n \frac{\K_{i-1}}{\N})\text{\space ,}
\end{equation}
with the difference that we subtract the overlap and need to compute $\sum \K_i$. 
The number of weights not covered on average by any ticket is then $\N - \sum \K_i$.

Analogous to the previous section, we approximate the spread 
by computing the maximal intersection and use this value to 
estimate the variance.
After \cite{pukelsheim1994three}, $95$\% of any uni-modal distribution is entailed in $3 \times \sigma$. We hence consider the largest overlap as
\begin{equation}
\K_i^{\max} = \n -  \n \frac{\K_{i-1}^{\max}}{\N} + 3 \n \frac{\N-\K_{i-1}^{\max}}{\N}\frac{\N - \n}{\N -1}\text{\space .}
\end{equation}
and obtain the largest covered weights as $\sum \K_i^{\max}$. 
The difference between these two terms leads to an approximation of the real variance.
To preserve the interpretation of the $95$\% interval, we plot the distribution with mean and set the derived variance as $2\sigma$ of a normal distribution.

\section{Output similarities between winning tickets.}\label{sec:ckaFull}
We here depict the full results from Section~\ref{sec:weightSpaceSymmetry} 
in the main paper, e.g. CKA similarity and $L_2$ distances for Fashion MNIST, CIFAR10 and the ResNet on CIFAR10.

\begin{figure}
    \includegraphics[width=0.49\textwidth]{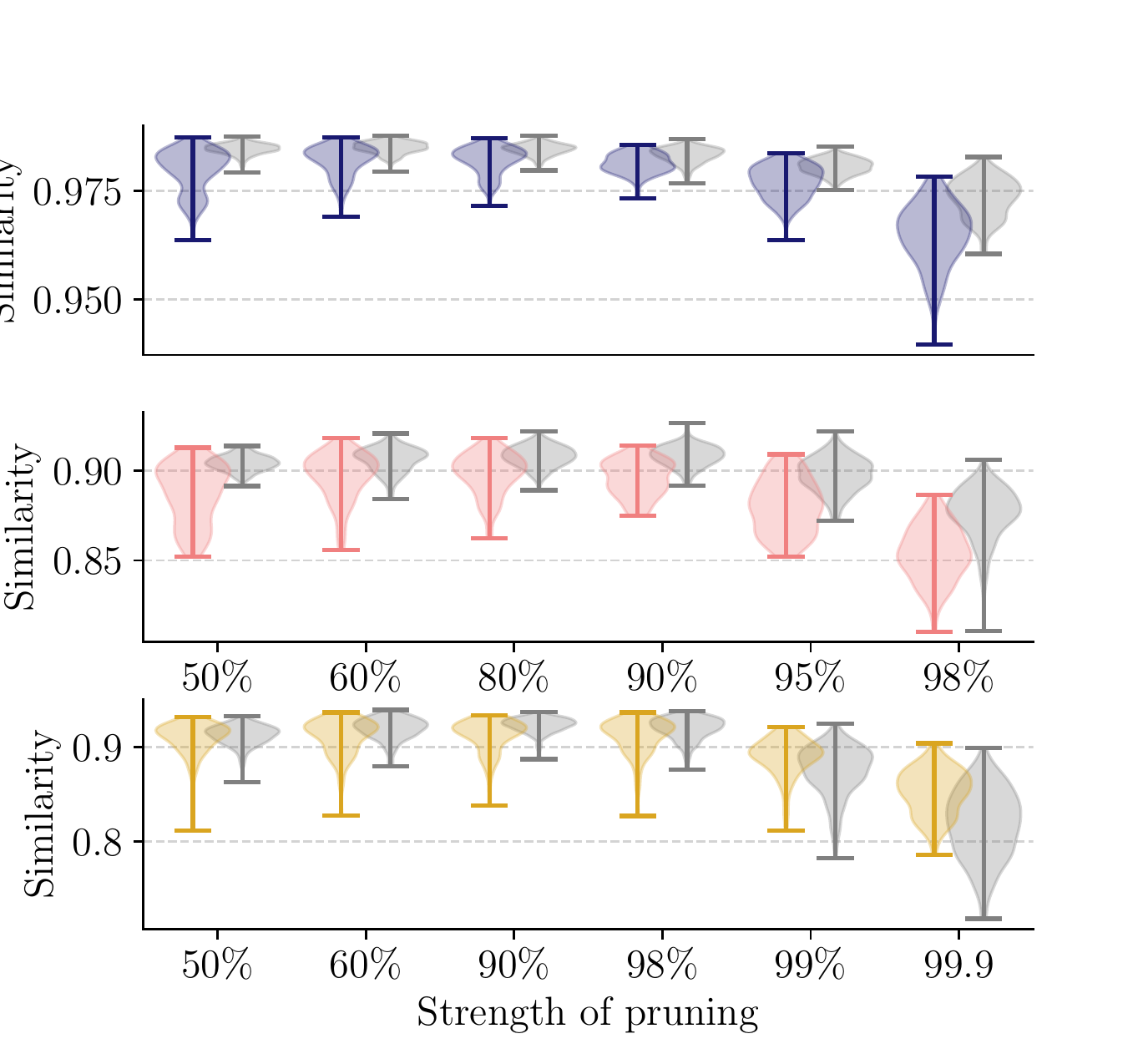}
    \includegraphics[width=0.49\textwidth]{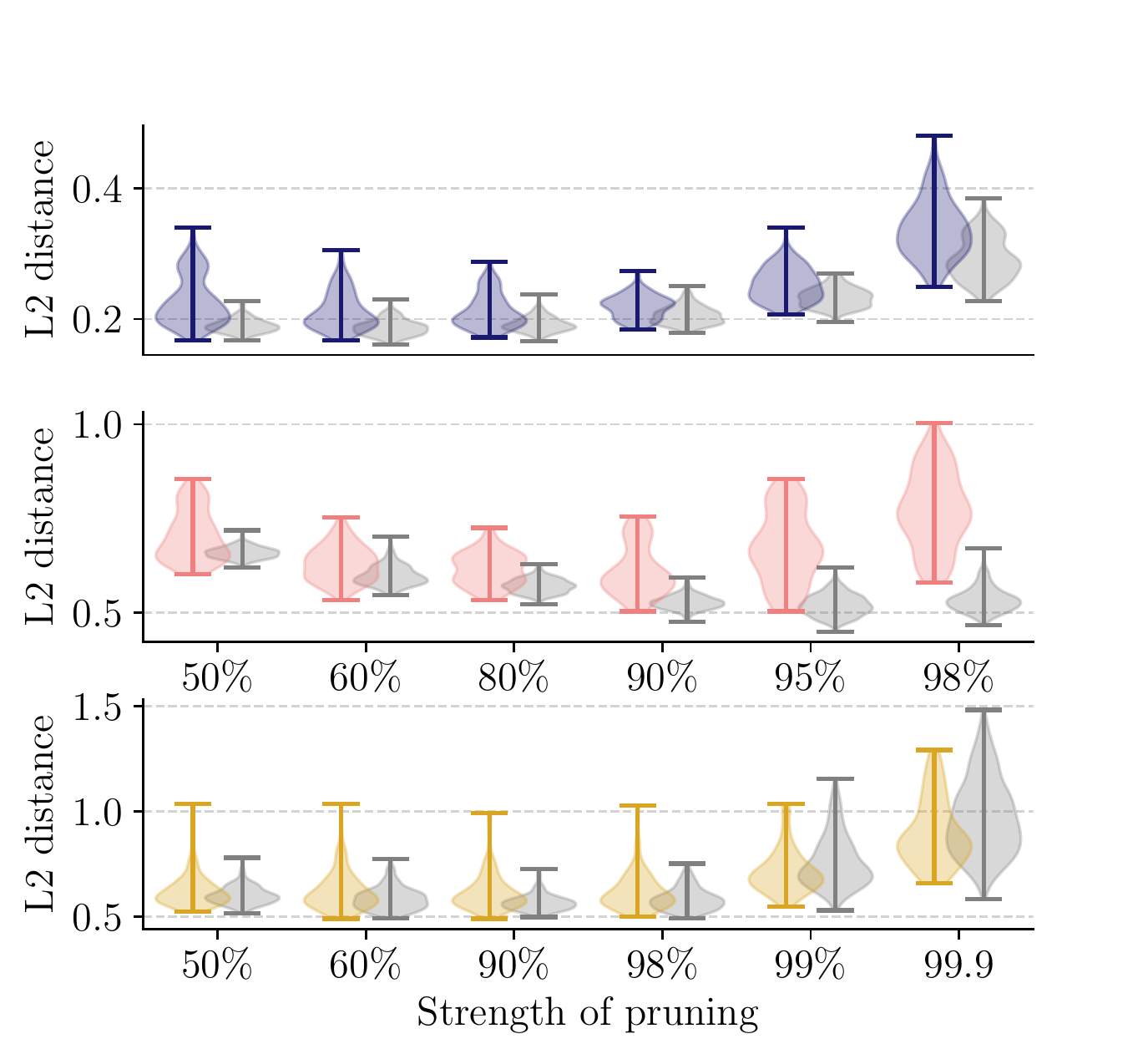}
  \caption{CKA similarity (left) and $L_2$-distance (right) between winning tickets within (colored) and across (gray) seeds for MNIST, CIFAR, and ResNet on CIFAR (from top to bottom).}
  \label{fig:CKAL2}
\end{figure}

\textbf{Results.} The tickets do not exist due to the weight space symmetry: no two tickets yield the exact same output.
 Overall, Fashion MNIST outputs across tickets are most similar. Second highest are ResNet outputs on CIFAR outputs, and least similar the small networks on CIFAR. In general, the similarity remains stable and decreases with pruning iteration $5$, somewhat similar to the accuracy. All networks are slightly more similar across seeds than within seeds. This holds for both metrics. 
In general, both measures show that the outputs diverge as pruning iterations increase.

\textbf{Conclusion.} Although the similarities between the outputs of the different pruned networks and winning tickets are similar, they are not equivalent.

\section{Details of fixed-randomness experiments}\label{sec:bonus}
In this Appendix, we discuss in more detail some of the experiments in Section~\ref{sec:fixed}.
All plots were generated with the randomness partially fixed.
We first analyze the overlap of winning tickets in detail.
We then discuss the amount of statistical significance of these overlaps,
before we discuss the overlaps across different tasks and when layers sizes are changed.

\subsection{Are winning tickets unique?}
\begin{figure*}[tb]
  \begin{subfigure}[b]{0.3\textwidth}
    \includegraphics[width=\textwidth]{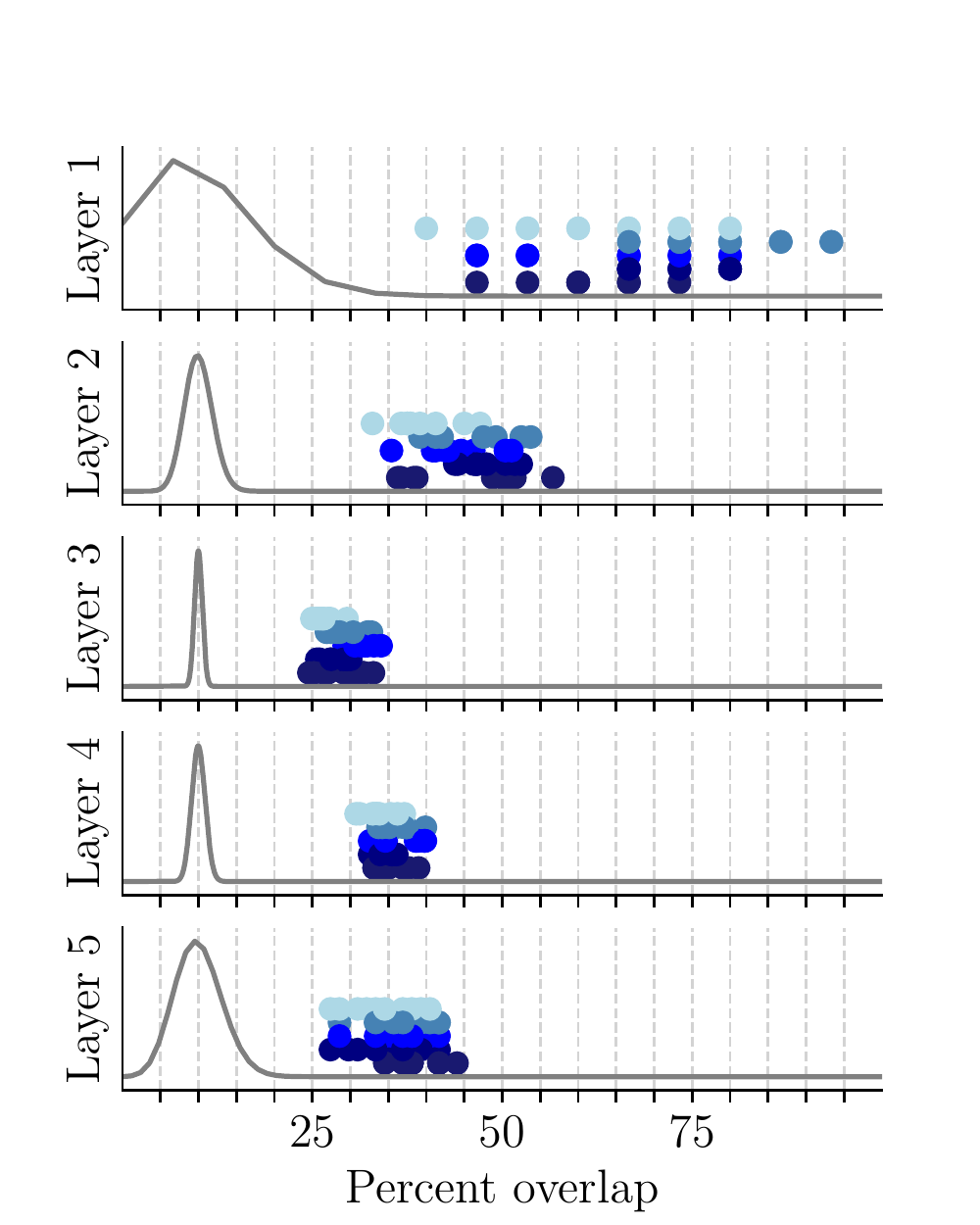}
    \caption{Fashion MNIST, $80$\% of weights pruned in $3$ iterations.}
    \label{fig:overlapLaterPruneMF}
  \end{subfigure}
    \hspace{7pt}
  \begin{subfigure}[b]{0.3\textwidth}
    \includegraphics[width=\textwidth]{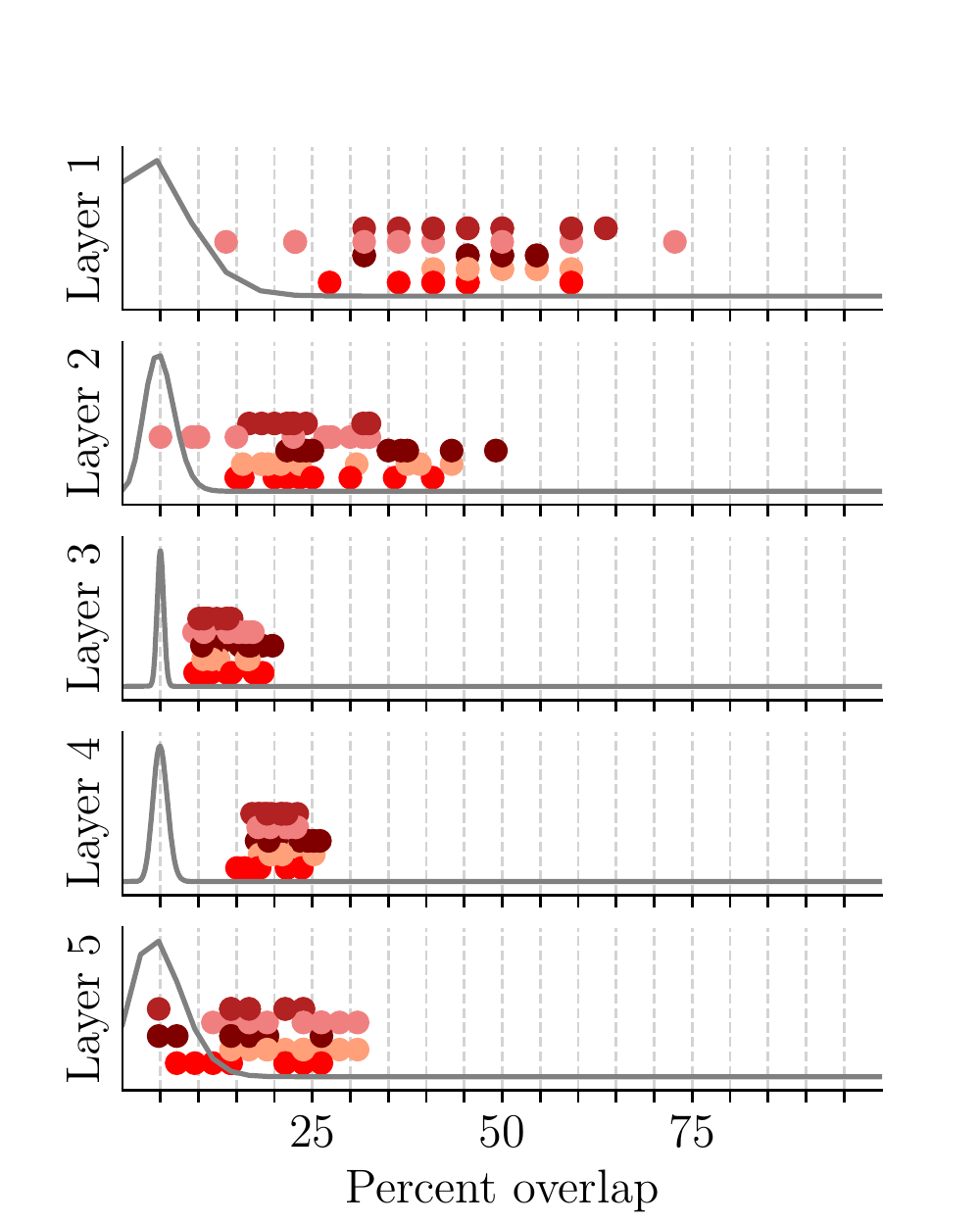}
    \caption{CIFAR10, $90$\% of weights pruned in $4$ iterations.}
    \label{fig:overlapLaterPruneCF}
  \end{subfigure} 
      \hspace{7pt}
    \begin{subfigure}[b]{0.3\textwidth}
    \includegraphics[width=\textwidth]{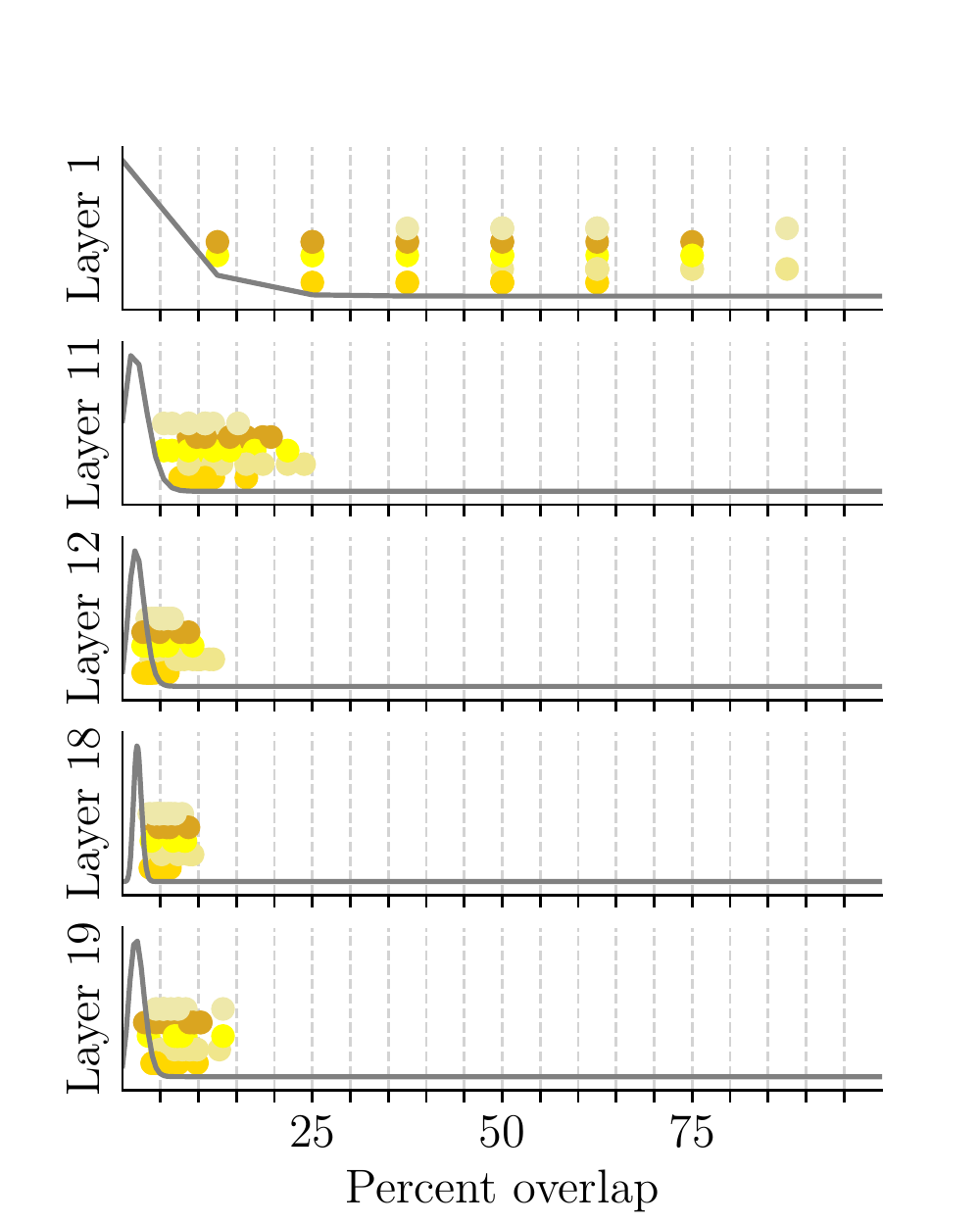}
    \caption{CIFAR10, $90$\% of weights pruned in $3$ iterations.}
    \label{fig:overlapLaterPruneRF}
  \end{subfigure} 
  \caption{Percentage of overlap between pruned masks (pruning after 15 epochs) of 5 runs, each using 5 initializations, with partially fixed randomness. Each initialization is one shade, $y$ position is based on seed and carries no further meaning. The gray curve is the expected overlap.}
  \label{fig:overlapLaterPruneF}
\end{figure*}

We compare the overlaps between all tickets generated from one initialization (case \textbf{A} from Figure~\ref{fig:setUP}) 
after iterative pruning with partially fixed randomness. Figure~\ref{fig:overlapLaterPruneF} shows pruning levels of $80$\% (Fashion MNIST), $90$\% (CIFAR) and $90$\% (ResNet, CIFAR). 

\textbf{Results.} On Fashion MNIST (Figure~\ref{fig:overlapLaterPruneMF}), the expected overlap lies around $10$\%.
The overlaps are generally less than $60$\%, but strongly biased towards the more-overlap side. 
The first layer shows very high overlaps, some of which are higher than $70$\%.
The second layer's overlaps range between $30$\% and $60$\%. 
The third layer's overlaps are scattered with low variance around $30$\%.
Analogously, the fourth layer's overlaps show little variance, and lie between $30$\% and $40$\%.
The last layer's overlaps vary between $25$\% and $45$\%.

On CIFAR (Figure~\ref{fig:overlapLaterPruneCF}), we investigate a smaller ticket, sized $10$\% of the original weights.
The expected overlaps vary around $5$\%.
In particular the first layer shows high overlaps, some of which are larger than $50$\%.
The second layer's overlaps show high variance and lie between $5$\% and around $45$\%.
The overlaps of the third layer exhibit low variance and range between $10$\% and $20$\%.
Analogously, the fourth layer's overlaps show little variance, are however scattered between $15$\% and $25$\%.
In the last layer, overlaps vary between $5$\% and $32.5$\%.

The expected overlaps of ResNet are around $2.5$\%, as we consider a small ticket of only $10$\% of the original weights. Overall, the overlaps are much lower with much lower variance, except the first layer, which exhibits overlaps between $12.5$\% and $92$\%.
The eleventh layers overlaps vary between $5$\% and $25$\%.
The later three plotted layers, namely twelve, eighteen and nineteen show overlaps between $2.5$\% and $10$\% (eighteenth layer) to $15$\% (twelfth and nineteenth layer).

\textbf{Conclusion.} 
Tickets for the same initialization are not equivalent.
Yet, the overlap is larger than expected  when randomness is partially fixed.
We investigate how many of these results are statistically significant in the following section.

\subsection{Statistical significance of overlaps for partially fixed randomness}\label{sec:significance}
\begin{figure}
  \includegraphics[width=\linewidth]{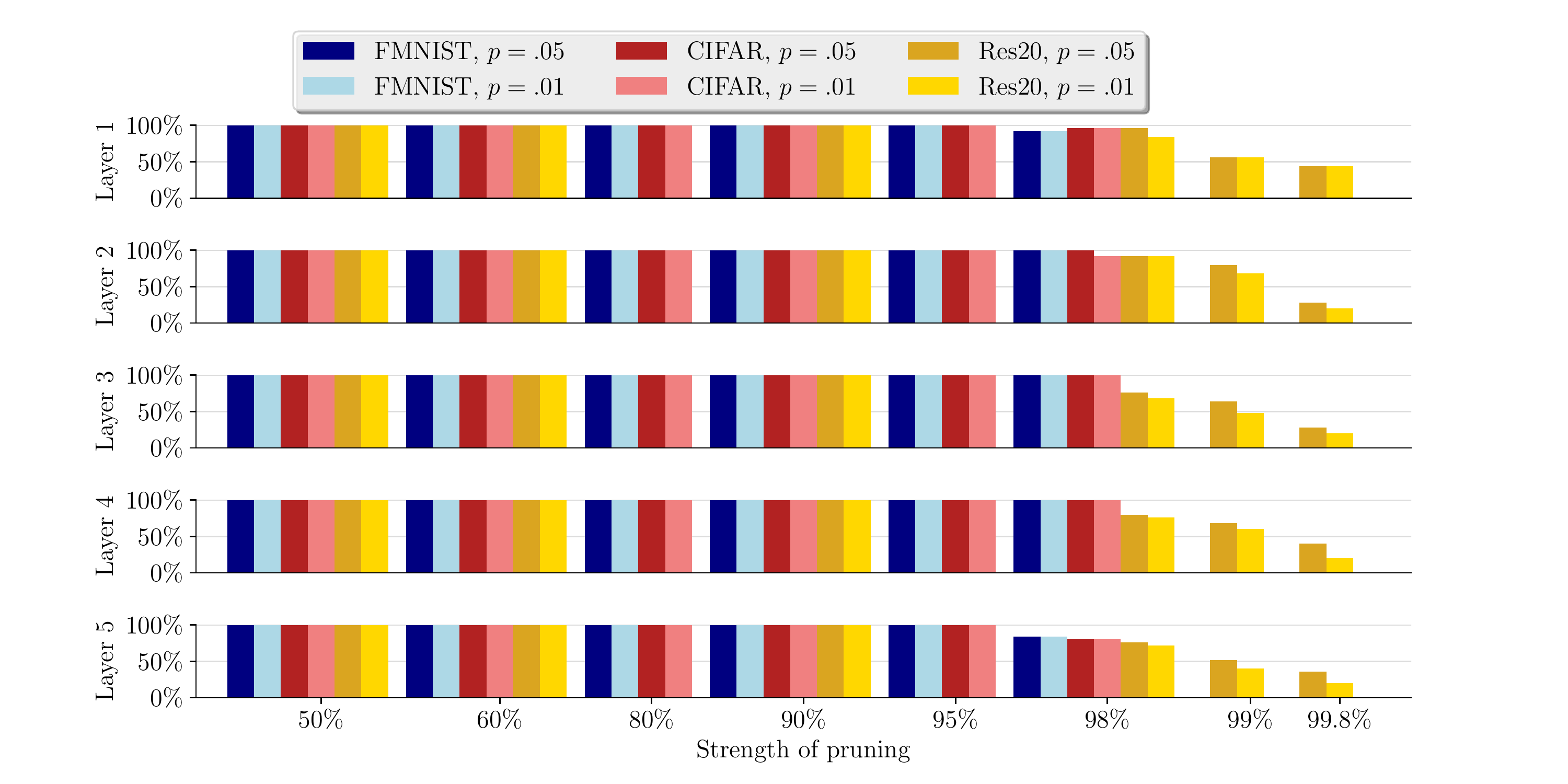}
  \caption{Computing percent of overlaps that are statistically significant in an iteration of pruning, with partially fixed randomness. Layer for ResNet are first, eleventh, twelfth, eighteenth, nineteenth.}
  \label{fig:significance}
\end{figure}

To determine how unusual the overlaps with partially fixed randomness are, we test their statistical significance given the hyper-geometric.
To this end,  we compute the $95$\% and $99$\% two sided interval of the hyper-geometric distribution.
Afterwards, we test how many of the observed similarities lie outside this range.
We plot the percentage of statistically significant similarities in Figure~\ref{fig:significance}.

\textbf{Results.}  The fraction of statistically significant values is almost always around $100$\%, with little difference depending on the $p$-value. 
However, the percentages decreases in the first and last layers for higher levels of pruning and for higher levels of pruning in general for ResNet.

On Fashion MNIST, the statistical significance is $100$\% for all layers in the first, second and third pruning iteration. 
The first layer shows a slight decrease starting from the sixth iteration.  
The second, third and fourth layer show no decrease in the amount significant overlaps.
In the last layer, significance decreases starting from our last or sixth pruning step. 

On CIFAR, the observations are analogous. On ResNet, the results are more pronounced, where the amount of significant overlaps decreases consistently from step four.
In step four, the percentage is however still high, and above $75$\%. 
In the next pruning step it decreases and lies around and above $50$\%. 
In the last step, the amount of statically significant high overlaps is below $50$\%. 
 
\textbf{Conclusion.}
Most observed overlaps are unusual when we underlay the hyper-geometric distribution. 
Yet, towards more iterations in pruning, the amount of statistically significant values decreases.

\subsection{Rank correlation of tickets with partially fixed randomness}
We repeat the computation of rank correlation for partially fixed randomness.
As before, a high correlation of $1.0$ implies that the order is preserved, $-1.0$ means the order is inverted, $0$ that there is no relationship in terms of rank correlation. If this correlation is positive, this supports that the final weights are correlated with the initial weights' magnitude.
We plot pruning levels of $80$\% (Fashion MNIST), $90$\% (CIFAR) and $90$\% (ResNet, CIFAR). 
The correlations are depicted in Figure~\ref{fig:corrsLaterPrune}. 

\textbf{Results.} On Fashion MNIST (Figure~\ref{fig:corrsLaterPruneM}), the correlations are all above $0.65$. 
More specifically, the first layer's correlations vary around $0.8$ up to $1.0$, the second layer's ranges between $0.75$ and $0.85$.
The third layer exhibits correlations between $0.7$ and $1.0$.
The fourth layer's correlations range from $0.9$ till $1.0$, with an outlier below $0.8$.
The last layer's correlation lie between $0.65$ and $0.95$.

\begin{figure*}[tb]
  \begin{subfigure}[b]{0.3\textwidth}
    \includegraphics[width=\textwidth]{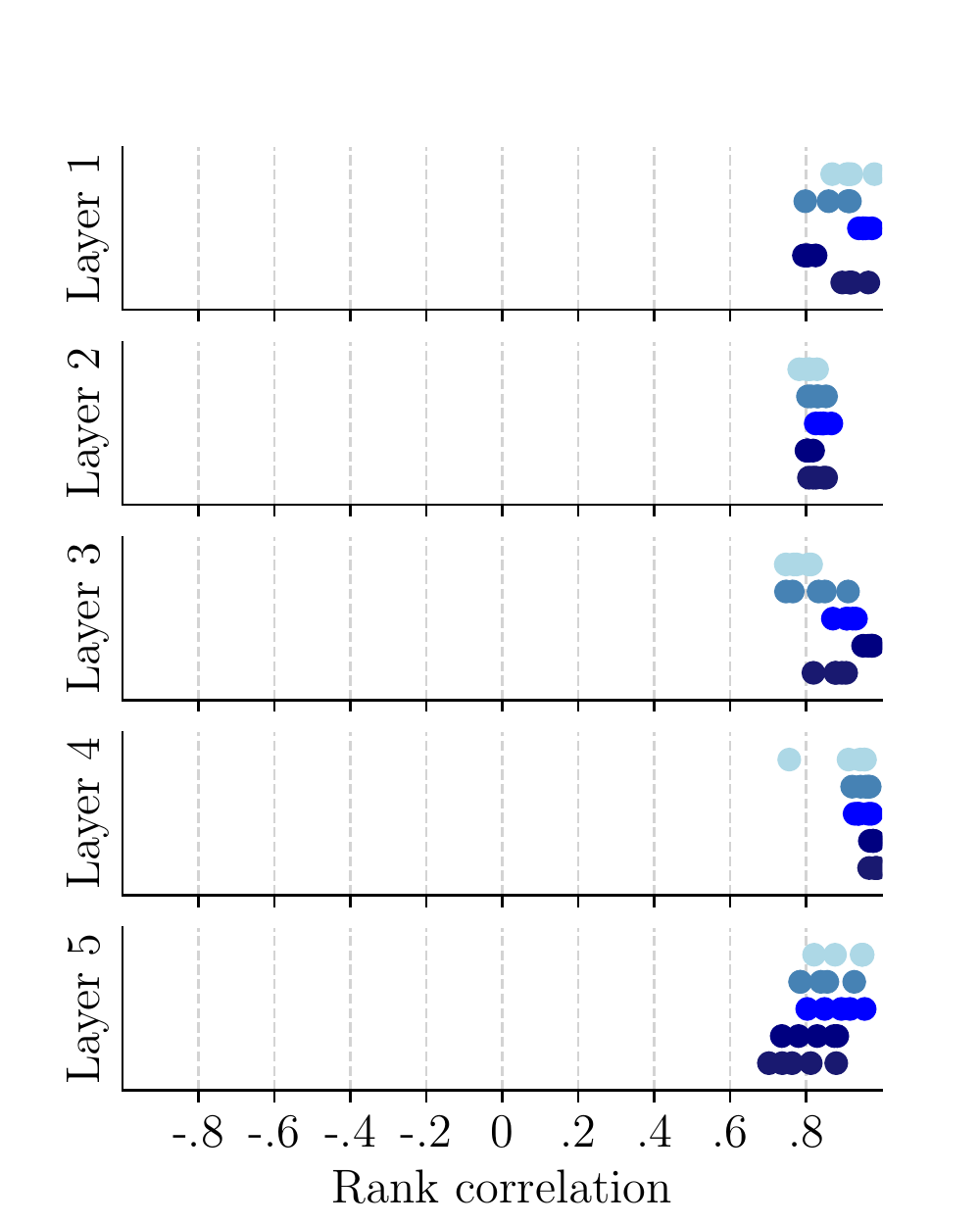}
    \caption{Fashion MNIST, $80$\% of weights pruned in $3$ iteration.}
    \label{fig:corrsLaterPruneMF}
  \end{subfigure}
    \hspace{7pt}
  \begin{subfigure}[b]{0.3\textwidth}
    \includegraphics[width=\textwidth]{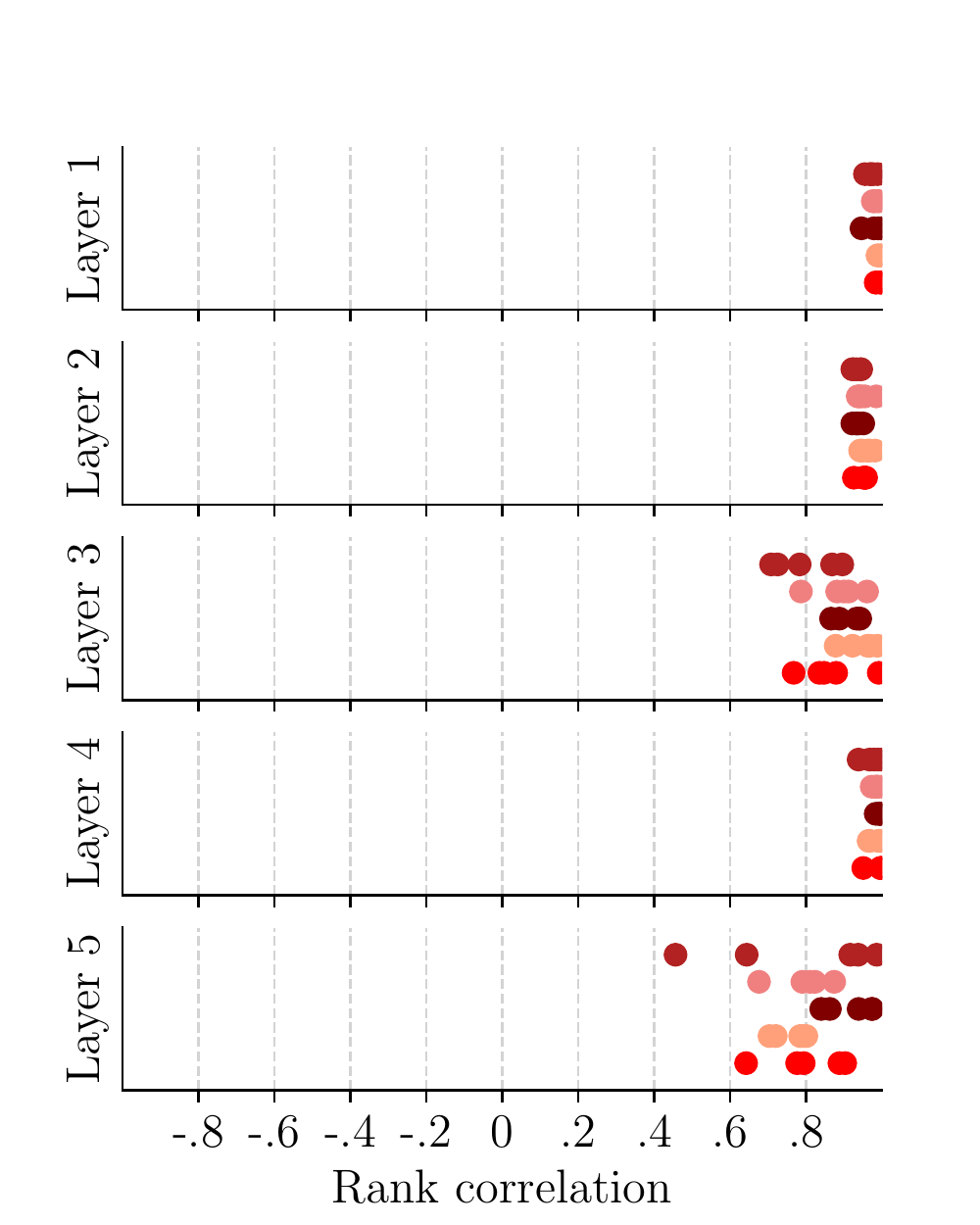}
    \caption{CIFAR10, $90$\% of weights pruned in $4$ iterations.}
    \label{fig:corrsLaterPruneCF}
  \end{subfigure} 
      \hspace{7pt}
    \begin{subfigure}[b]{0.3\textwidth}
    \includegraphics[width=\textwidth]{figures/NRCIFARResNet3Corrs.pdf}
    \caption{CIFAR10, $90$\% of weights pruned in $3$ iterations.}
    \label{fig:corrsLaterPruneRF}
  \end{subfigure} 
  \caption{Rank correlation between initial weights and weights in masks (pruning after 15 epochs), each using 5 initializations, with partially fixed randomness. Each initialization is one color shade, $y$ position is based on seed and carries no further meaning.}
  \label{fig:corrsLaterPruneF}
\end{figure*}

On CIFAR (Figure~\ref{fig:corrsLaterPruneC}), the ticket contains $10$\% of the original weights. As before, the correlations are high.
 The first layers correlations are very high, and lie between $0.95$ and $1.0$.
 In the second layer, values are slightly lower, varying between $0.9$ and $1.0$.
 The third layers values are between $0.7$ and $1.0$. 
 The fourth and last layer's correlations range between $0.9$ and $1.0$.
The last layer's correlation show the largest variance, with values ranging from $0.6$ to $1.0$, 
and an outlier around $0.4$.  

The ResNet's correlations are overall much lower, and lie around $0.1$.  
The first layer's correlations are rather high and range between $0.2$ and $0.4$.
The eighth, eleventh and twelfth layers' correlations all lie between $0$ and $0.2$.
The last layer shows very high variance of the correlations.
There is one cluster between $0$ and $0.15$, two small ones around $0.35$ and $0.55$ and one close to $1.0$.

\textbf{Conclusion} 
The correlations, albeit much less pronounced for the ResNet, are all positive, showing that the initial weights have a large influence on the resulting ticket.

\subsection{Data and winning tickets for fixed randomness}
\begin{wrapfigure}{r}{0.63\textwidth}
  \vspace{-15pt}
  \begin{subfigure}[b]{0.31\textwidth}
    \includegraphics[width=\textwidth]{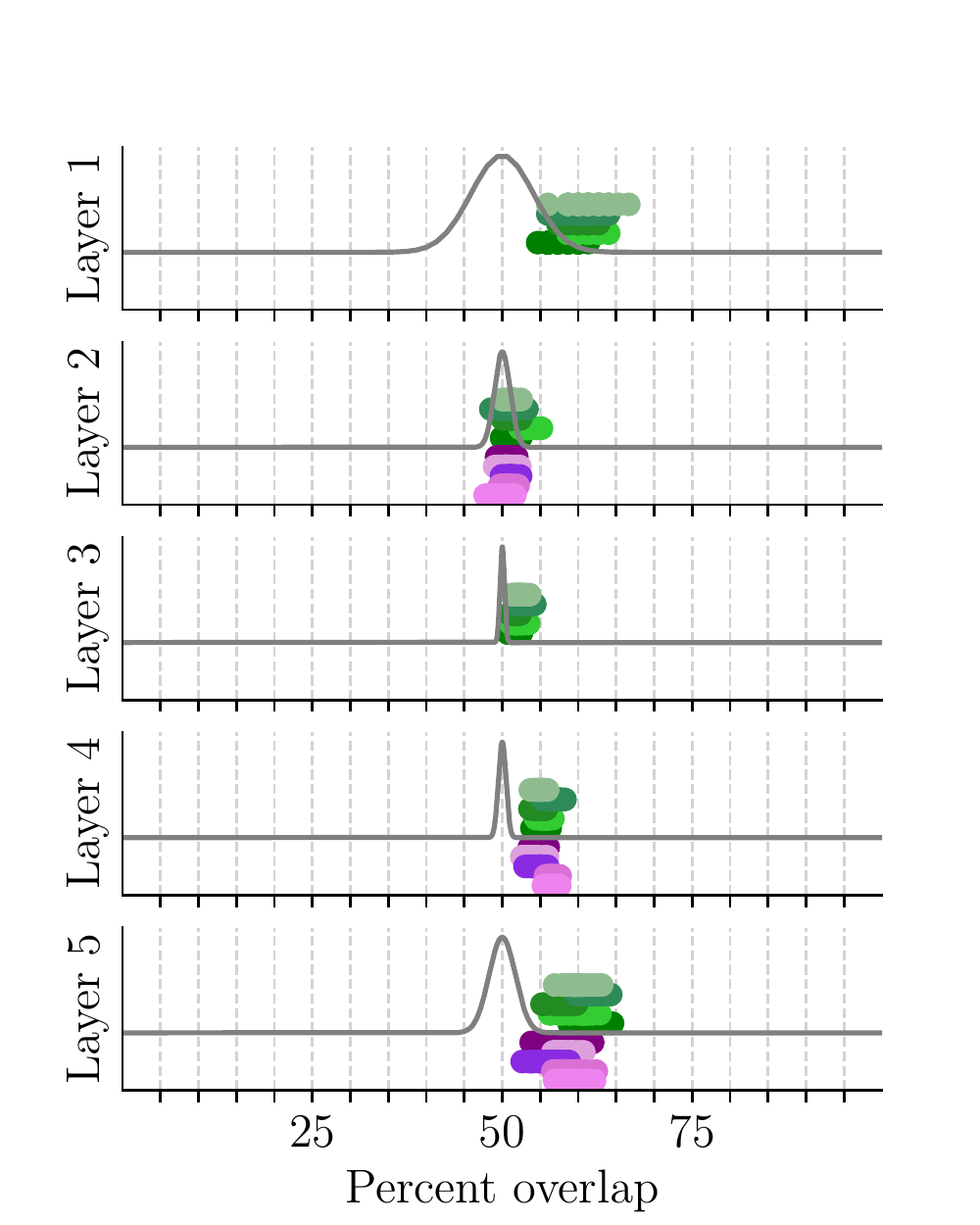}
    \caption{$50$\% of weights pruned in one iteration.}
  \end{subfigure}
   \begin{subfigure}[b]{0.31\textwidth}
    \includegraphics[width=\textwidth]{figures/4AcrossDataFix.pdf}
    \caption{$90$\% of weights pruned in four iterations.}
  \end{subfigure}
  \caption{Comparing pruned masks from different datasets and pruning iterations. Percentage of overlap between pruned masks (pruning after 15 epochs) of 5 runs, across 5 initializations, with partially fixed randomness. Differences between MNIST and Fashion MNIST are shown in green, between CIFAR and Fashion MNIST in purple.}
  \label{fig:dataOverlap}
    \vspace{-15pt}
\end{wrapfigure}

We now investigate if fixed randomness leads to similar tickets even across datasets.
As the CIFAR networks, due to the larger dimensionality of the dataset, cannot have all layers same-sized, we additionally train networks of same architecture on the MNIST dataset~\cite{lecun1998gradient}. 
In Figure~\ref{fig:dataOverlap},
we measure the distance from Fashion MNIST to MNIST tickets (green) among all layers. In the same figure,
the distances from Fashion MNIST to CIFAR10 masks (purple) are only measured on layers of the same size (two, four, and five).

\textbf{Results.} The overlaps are generally higher than chance. 
Yet, they are less pronounced than tickets of the same dataset with partially fixed randomness. 
There are no significant differences between the overlaps with CIFAR or MNIST, although the architecture of the CIFAR networks differs in two layers from the MNIST networks.

In the first pruning iteration,
the first layer's overlaps are slightly higher than expected, and range between $52$\% and $65$\%. 
In the second layer, the overlaps are as expected, with a slight bias towards more overlap.
The third layer exhibits overlaps between $50$\% and $55$\%, hence larger than expected.
On the fourth layer, the overlaps range between $52.5$\% and $65$\%. 
The last layer, as the previous, shows higher than expected overlap ranging between $52.5$\% and $65$\%.
In the fifth iteration, the overlaps exhibit up to $30$\% overlap  in the first layer.
The second layer shows only a slight bias towards more overlap.
The third layer shows higher overlaps than expected, ranging between $5$ and $10$\%.
Also the fourth layer exhibits higher overlaps than expected, between $7.5$ and $16$\%.
The last layer, in contrast to the first pruning step, shows only few unexpectedly high overlaps.

\begin{wrapfigure}{r}{0.62\textwidth}
  \vspace{-20pt}
  \begin{subfigure}[b]{0.3\textwidth}
    \includegraphics[width=\textwidth]{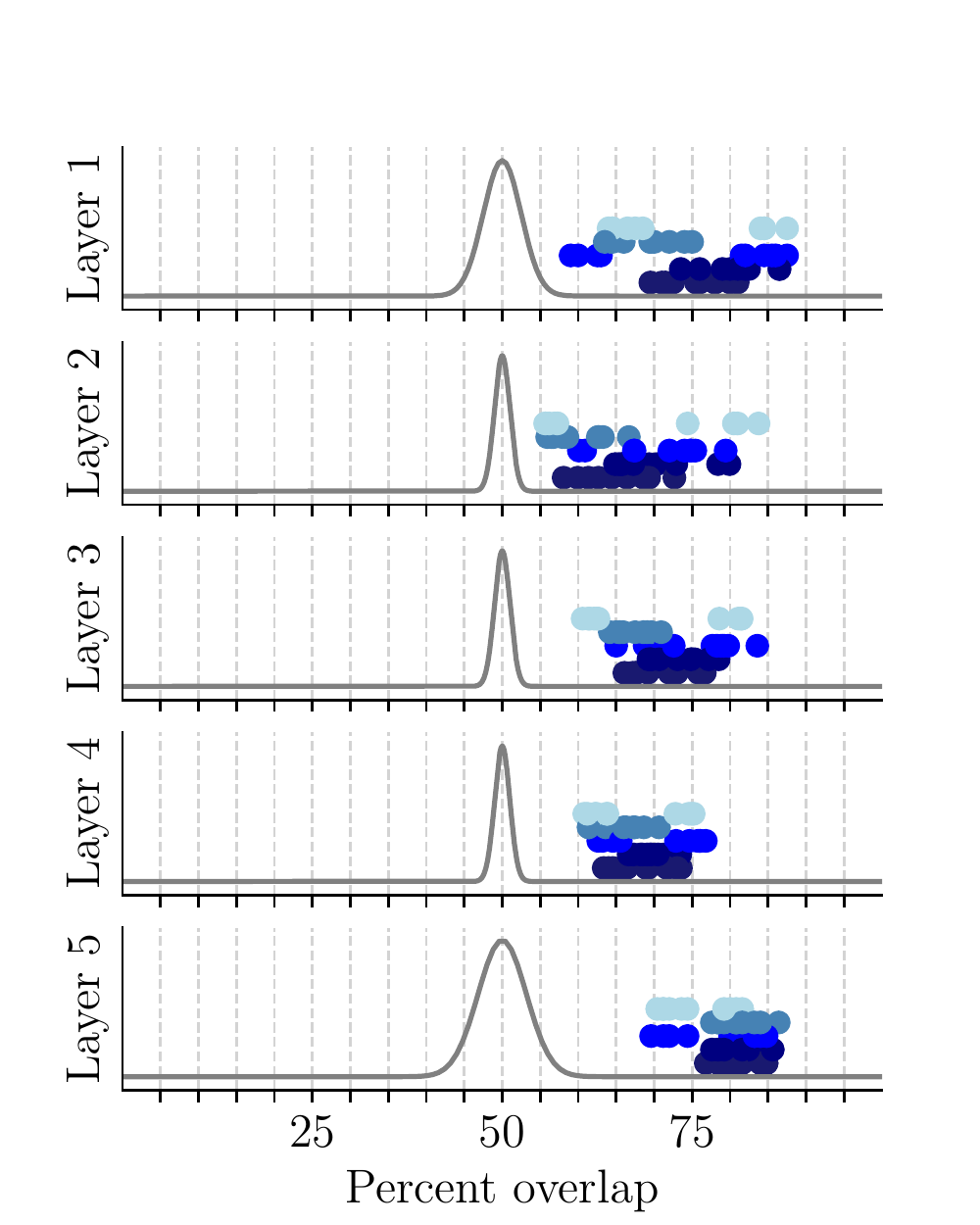}
    \caption{Fashion MNIST, $50$\% of weights pruned in one iteration.}
  \end{subfigure}
  \begin{subfigure}[b]{0.3\textwidth}
    \includegraphics[width=\textwidth]{figures/4OverlapSMFIX.pdf}
    \caption{Fashion MNIST, $90$\% of weights pruned in four iterations.}
  \end{subfigure} 
  \caption{Percentage of overlap between pruned masks (pruning after 15 epochs) of 5 runs, across 5 initializations, with partially fixed randomness. Each initialization is one blue shade. The gray curve is the expected overlap. We adjusted the second, third, and fourth layer to be of the same size.}
  \label{fig:bonus}
    \vspace{-25pt}
\end{wrapfigure} 
\textbf{Conclusion.} The overlaps are in some cases higher than chance when training the networks on different tasks, given that the initial weights fixed and randomness is partially fixed. 

\subsection{The effect of layer size and position for fixed randomness}

In the previous Figures, the small networks' inner layers (here three and four) exhibited different overlaps than the outer layers. 
To investigate whether this is an effect of the layer size or its position, we train additional Fashion MNIST networks where all inner layers have approximately the same size (e.g., $2400$ or $2500$ weights, first layer $400$, last layer $250$). 
We plot the resulting overlaps in Figure~\ref{fig:bonus}.

\textbf{Results.} Layer size has more effect on the results than position in the network. When inner layers are of similar size, their pruning overlaps look more similar to outer layers. 
For a low pruning level ($50$\%), the spread of the overlaps in the first layer increases. The values now range from $57$\% to $88$\%.
The variance of overlaps in the second layer also increased, with values between $55$\% and $80$\%.
The third and fourth layer are similar, and overlaps range between $65$\% and $82$\% (third) and $77$\% (fourth).
The last layer's overlaps now range between $70$\% and $90$\%.

At a higher pruning level ($95$\%), the spread of the first layer decreases slightly. 
The values now range between $5$\% and $40$\%.
Also the following layer's variance decreases.
The second show overlaps from $5$\% to $20$\%.
Third and fourth layer are similar, the overlaps range in general between $10$\% and $25$\% (third) and $22$\% (fourth). 
The last layer varies much more, overlaps now lie between $0$ and $42$\%.

\textbf{Conclusion.} The layer's position in the networks has no influence on the overlaps. 
There is also no clear correlation between layer size and variance of the overlaps. 

\subsection{Weight-space symmetry for partially fixed randomness}
\begin{wrapfigure}{l}{0.62\textwidth}
  \vspace{-15pt}
    \includegraphics[width=0.6\textwidth , trim={0 0 0 1cm},clip]{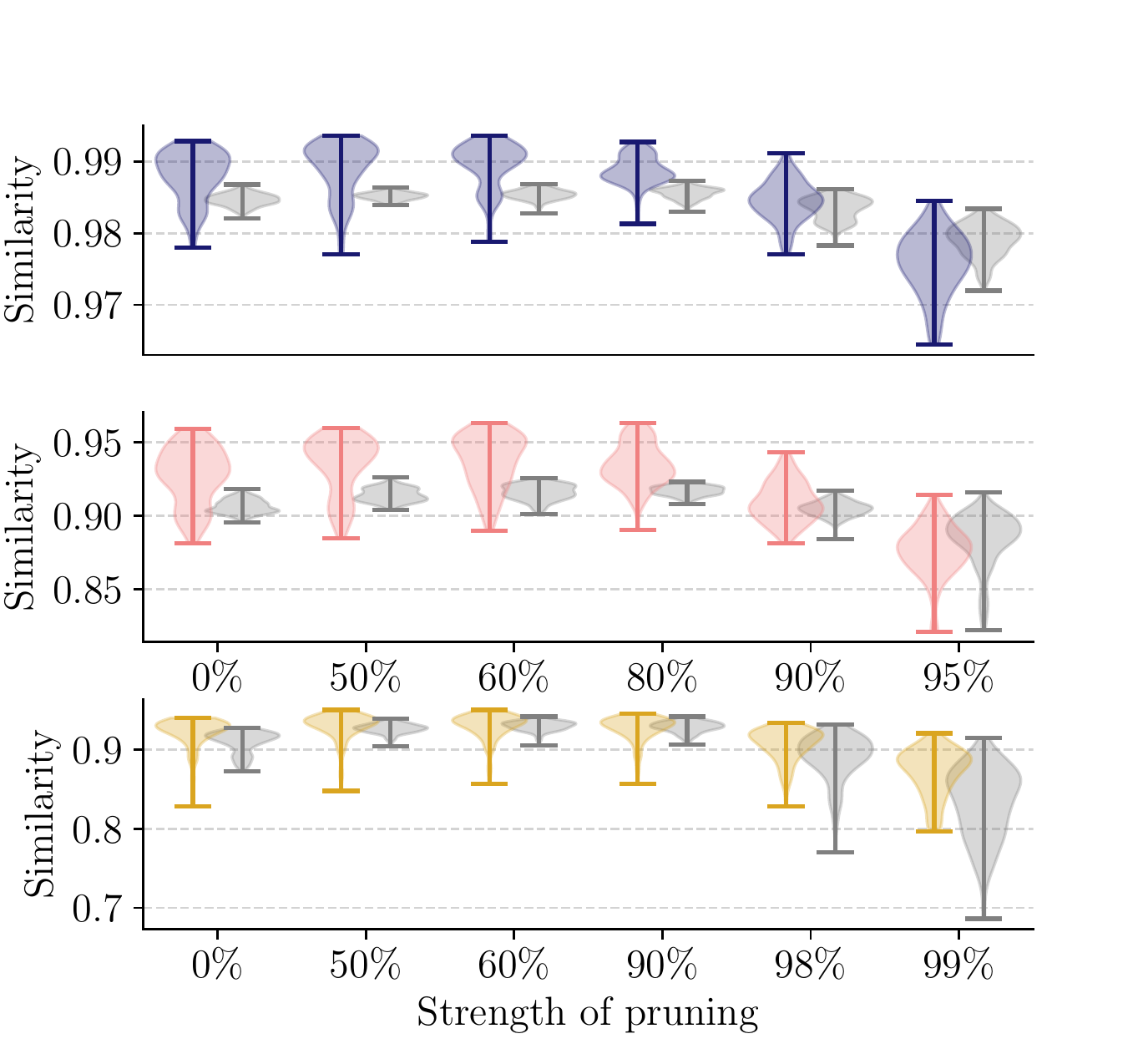}
  \caption{CKA similarity between winning tickets with partially fixed randomness within (colored) and across (gray) seeds for MNIST, CIFAR, and ResNet on CIFAR (from top to bottom). Plot is best seen in color.}
  \label{fig:NRCKA}
    \vspace{-15pt}
\end{wrapfigure}
As before, we investigate whether the tickets (this time derived with partially fixed randomness) are in fact instances of the same network. As tickets now exhibit more overlap, they might be more similar concerning their weights, too.  As both $L_2$ and CKA measure are similar, we drop the $L_2$ based results and only show CKA in Figure~\ref{fig:NRCKA}. The colored violin diagrams show the CKA measure between networks derived from the same initialization (compare \textbf{A} in Figure~\ref{fig:setUP}), the gray violins the similarities across initializations (compare \textbf{B} in Figure~\ref{fig:setUP}).

\textbf{Results.} The tickets do not exist due to the weight space symmetry. As before, no two tickets yield the exact same input. Overall, the CKA similarities are highest for Fashion MNIST, second highest for CIFAR, and lowest for ResNet on CIFAR. In general, the similarity decreases as the pruning level increases, somewhat similar to the accuracy. For the small networks on Fashion MNIST and CIFAR, the similarities across seeds are lower than within seeds. For the ResNet, they seem to be more similar, but at lower pruning levels, there are outliers showing less similar outputs. 

\textbf{Conclusion.} Although the similarities between the outputs of the different pruned networks and winning tickets are similar, they are not equivalent in the weight space. Compared to the setting where randomness is not fixed, the similarities between tickets derived from the same initial weights is now higher than across seeds. This is expected, as the networks are, due to the fixed randomness, not able to diverge as much as with full randomness.

\end{document}